\documentclass[sigconf]{acmart}
\usepackage{multirow}
\usepackage{stfloats}
\usepackage{subcaption}
\usepackage[table,xcdraw]{xcolor}
\AtBeginDocument{%
  \providecommand\BibTeX{{%
    \normalfont B\kern-0.5em{\scshape i\kern-0.25em b}\kern-0.8em\TeX}}}

\newcommand{\ie}{\textit{i.e., }}
\newcommand{\etal}{\textit{et al. }}
\newcommand{\eg}{\textit{e.g., }}

\newcommand{\ourmethod}{SPAN}
\newcommand{\modulator}{Multimodal Modulator}

\copyrightyear{2025} 
\acmYear{2025} 
\setcopyright{acmlicensed}\acmConference[MM '25]{In Proceedings of the 33rd ACM International Conference on Multimedia}{October 27--31, 2025}{Dubin, Ireland}
\acmBooktitle{Proceedings of the 33rd ACM International Conference on
Multimedia (MM '25), October 27-31, 2025, Dubin, Ireland}
\acmDOI{https://doi.org/10.1145/3746027.3755104}
\acmISBN{978-1-4503-XXXX-X/2025/06}

\begin{document}

\title{SPAN: Continuous Modeling of Suspicion Progression for Temporal Intention Localization}

\author{Xinyi Hu}
\orcid{0009-0002-2252-8589}
\authornote{Xinyi Hu and Yuran Wang are co-first authors.}
\affiliation{%
  \institution{National Engineering Research Center for Multimedia Software, Institute of Artificial Intelligence, School of Computer Science, \\Wuhan University}
  \institution{Hubei Key Laboratory of Multimedia and Network Communication Engineering}
  \city{Wuhan}
  \country{China}}
\email{huxywhu@whu.edu.cn}

\author{Yuran Wang}
\orcid{0000-0002-3065-2830}
\authornotemark[1]
\affiliation{%
  \institution{School of Mathematical Sciences, Peking University}
  \city{Beijing}
  \country{China}}
\affiliation{%
  \institution{National Engineering Research Center for Multimedia Software, Institute of Artificial Intelligence, School of Computer Science, \\Wuhan University}
  \institution{Hubei Key Laboratory of Multimedia and Network Communication Engineering}
  \city{Wuhan}
  \country{China}}
\email{yuranwang25@stu.pku.edu.cn}

\author{Ruixu Zhang}
\orcid{0009-0004-9985-0204}
\affiliation{%
  \institution{Tsinghua University}
  \city{Shenzhen}
  \state{Guangdong}
  \country{China}}
\affiliation{%
  \institution{National Engineering Research Center for Multimedia Software, Institute of Artificial Intelligence, School of Computer Science, \\Wuhan University}
  \institution{Hubei Key Laboratory of Multimedia and Network Communication Engineering}
  \city{Wuhan}
  \country{China}}
\email{zhangruixu@whu.edu.cn}

\author{Yue Li}
\orcid{0009-0004-0320-7178}
\affiliation{%
  \institution{National Engineering Research Center for Multimedia Software, Institute of Artificial Intelligence, School of Computer Science, \\Wuhan University}
  \institution{Hubei Key Laboratory of Multimedia and Network Communication Engineering}
  \city{Wuhan}
  \country{China}}
\email{lleeyue@163.com}

\author{Wenxuan Liu}
\orcid{0000-0002-4417-6628}
\affiliation{%
  \institution{School of Computer Science, Peking University}
  \institution{State Key Laboratory for Multimedia Information Processing, Peking University}
  \city{Beijing}
  \country{China}}
\email{lwxfight@126.com}

\author{Zheng Wang}
\orcid{0000-0003-3846-9157}
\authornote{Corresponding authors: Zheng Wang.}
\affiliation{%
  \institution{National Engineering Research Center for Multimedia Software, Institute of Artificial Intelligence, School of Computer Science, \\Wuhan University}
  \institution{Hubei Key Laboratory of Multimedia and Network Communication Engineering}
  \city{Wuhan}
  \country{China}}
\email{wangzwhu@whu.edu.cn}

\renewcommand{\shortauthors}{Xinyi Hu, Yuran Wang, Ruixu Zhang, Yue Li, Wenxuan Liu, Zheng Wang}

\begin{abstract}
Temporal Intention Localization (TIL) has emerged as a critical task in video surveillance, focusing on identifying varying levels of suspicious intention to enhance security monitoring. However, existing discrete classification methods fail to capture the continuous progression of suspicious intentions, significantly limiting both the early intervention capabilities and explainability of such systems. 
In this paper, we reconceptualize hidden intention modeling by shifting from discrete classification to continuous regression and propose Suspicion Progression Analysis Network (SPAN), which capture the fluctuations and progression of hidden intentions over time. 
Specifically, when analyzing the temporal progression of suspicion, we discover that suspicion exhibits long-term dependency and cumulative effects across extended sequences, characteristics significantly similar to the settings in Temporal Point Process (TPP) theory. Based on these insights, we formalize a suspicion score formula that effectively models continuous changes while accounting for these temporal characteristics. Furthermore, we propose Suspicion Coefficient Modulation to modulate suspicion coefficients in the suspicion formula by leveraging multimodal information reflecting different effects of suspicious actions. Notably, we introduce a Concept-Anchored Mapping method that quantifies associations between suspicious actions and predefined suspicious intention concepts, enabling understanding of not just actions occurring but also their potential underlying intentions.
Extensive experiments on the HAI dataset show that SPAN significantly outperforms existing methods, reducing MSE by 19.8\% and improving average mAP by 1.78\%,. Notably, SPAN achieves a 2.74\% mAP gain in low-frequency cases, indicating superior capability in capturing subtle behavioral changes. Compared to discrete classification systems, out continuous suspicion modeling method enables earlier detection and more proactive interventions, substantially enhancing both system explainability and practical utility in security applications.
\end{abstract}






\maketitle
\begin{figure}[t]
    \centering
    \includegraphics[width=\linewidth]{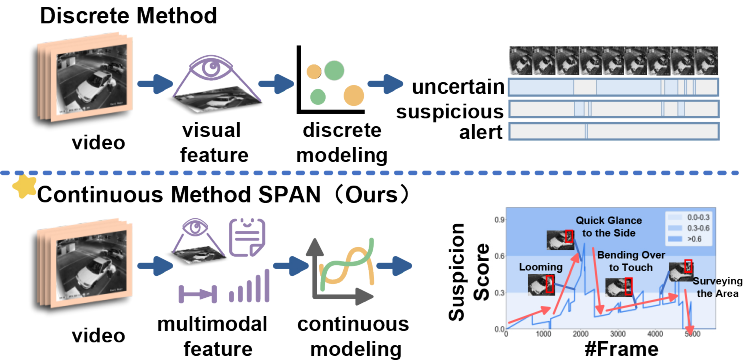} 
    \caption{Comparison between discrete and our continuous SPAN. The discrete method classifies suspicion into three fixed levels via visual features, while SPAN models suspicion as a continuous progression through multimodal features, enabling more precise, earlier detection of suspicious intentions with improved explainability}
    \label{fig:method_comparison}
\end{figure}

\section{Introduction}

Traditional intention recognition research~\cite{wei2017inferring, wei2018and, liu2020spatiotemporal, fang2021learning, xu2021trajectory, xu2022gaze, 10418196} primarily focuses on explicitly expressed \emph{normal intention}, limiting its capacity to detect hidden intentions prior to abnormal actions.
To address this limitation, \citeauthor{zhou2023uncovering}~\cite{zhou2023uncovering} introduces the Hidden Intention Discovery task, which targets the identification of \emph{hidden intention}, a concealed motive or goal preceding abnormal actions. For fine-grained recognition, \citeauthor{qi2024predicting}~\cite{qi2024predicting} proposes the Temporal Intention Localization (TIL) task, which localizes pre-abnormal actions associated with hidden intentions. They categorize suspicion into three discrete levels: uncertain, suspicious, and alert, establishing a baseline for pre-abnormal intention analysis.

However, the method in~\cite{qi2024predicting} produces discrete suspicion levels, which creates a fundamental gap for real-world monitoring. This method relies on fixed thresholds that trigger alerts only after reaching specific levels and lacks explanatory context during state transitions, limiting interpretability for security personnel.
In real-world scenarios, suspicious intentions evolve gradually through subtle behavioral cues, with experts identifying patterns rather than fixed states.

To address this gap, we reconceptualize hidden intention modeling by shifting from discrete classification to \textbf{continuous regression}. Continuous modeling offers three main advantages: it \textbf{detects rising trends before suspicion reaches critical thresholds}, enabling earlier intervention; it \textbf{captures the continuous progression of suspicion} through preemptive indicators interpretable as trajectories rather than static points; and it \textbf{enables dynamic threshold calibration} based on context, mitigating the limitations of hypersensitive or delayed static systems.

\begin{figure}[t]       
    \centering           
    \begin{subfigure}{0.45\linewidth}  
        \centering
        \includegraphics[width=\linewidth]{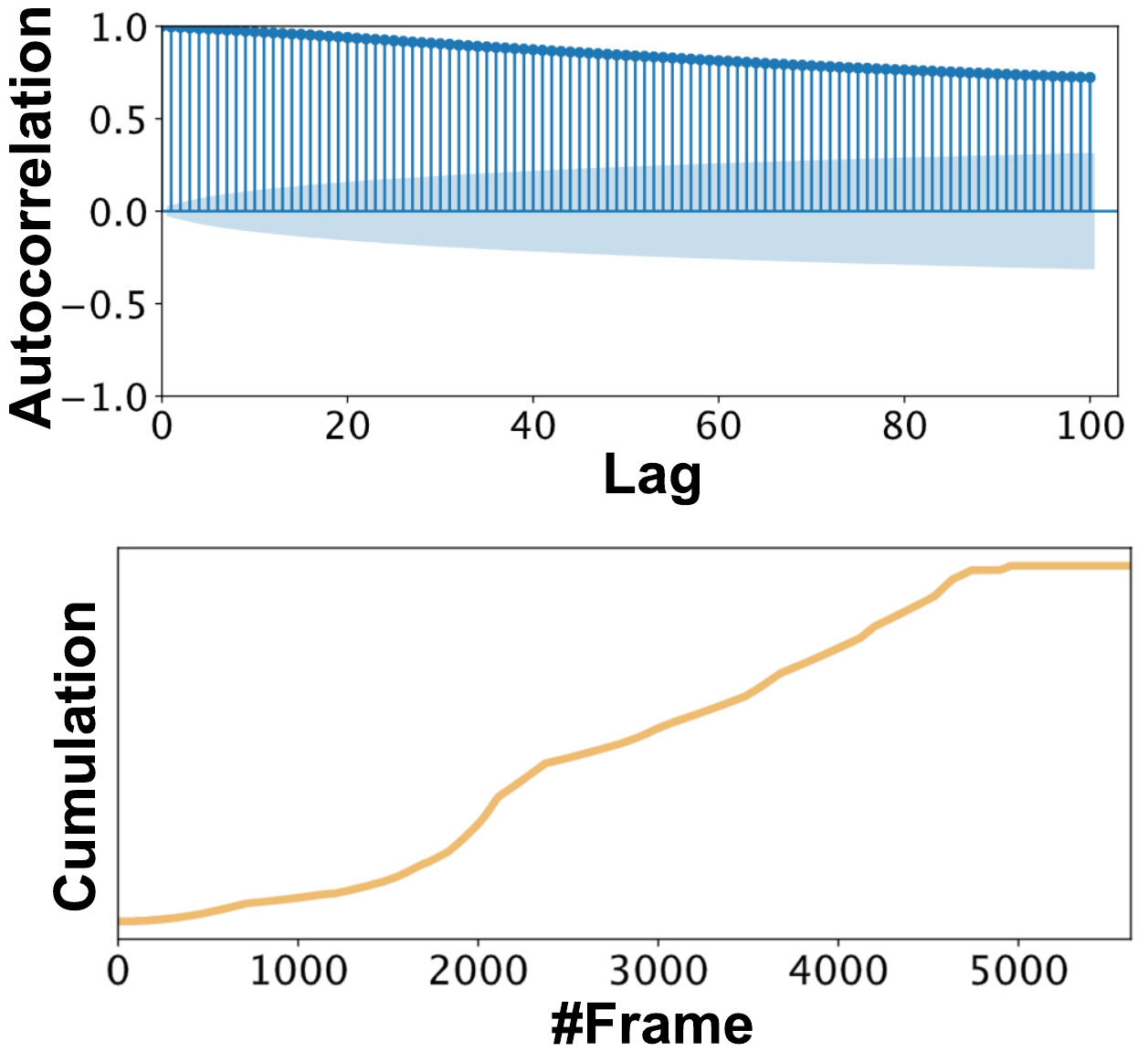}
        \caption{}
        \label{fig:motivation1}
    \end{subfigure}
    \hfill                
    \begin{subfigure}{0.45\linewidth}  
        \centering
        \includegraphics[width=\linewidth]{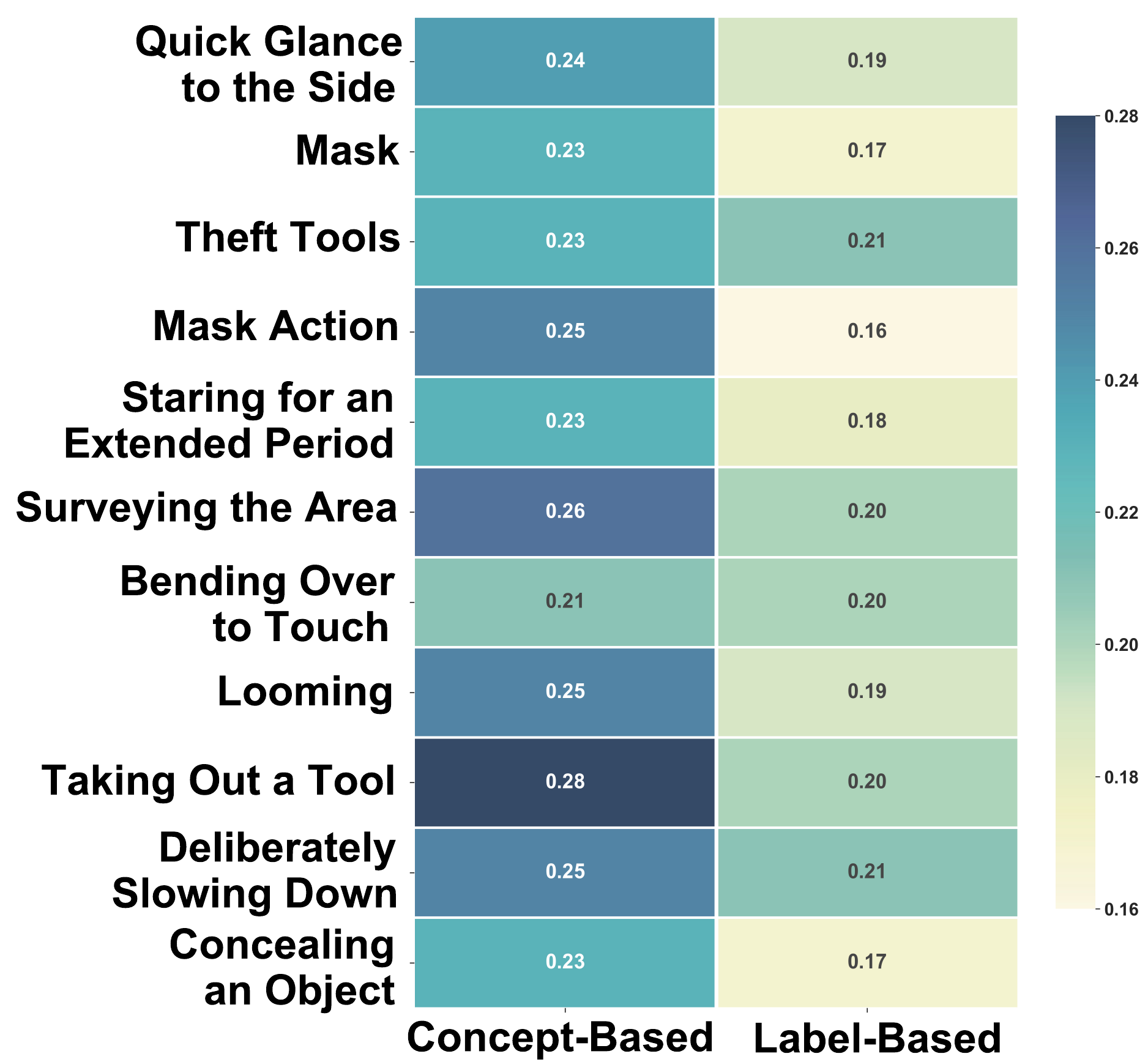}
        \caption{}
        \label{fig:motivation2}
    \end{subfigure}
    \caption{Motivation for our approach: (a) Temporal dependencies and cumulative effects in suspicion scores, (b) Comparison of textual and visual feature similarities.}
    \label{fig:motivation}
\end{figure}

However, reframing the problem from classification to regression introduces challenges in modeling and optimization. Continuous value prediction demands greater precision, necessitating the capture of subtle data patterns, while prediction errors are more sensitive to outliers.
To overcome this challenge, we analyze the temporal progression of suspicion score in Fig.~\ref{fig:motivation}(a), discovering that suspicion exhibits long-term dependency and cumulative effects across extended sequences, which are two key characteristics of suspicion progression, and these temporal characteristics significantly similar to the settings in Temporal Point Process (TPP) theory.
Moreover, we explore to bridge observable actions and hidden intentions by leveraging richer semantic descriptions, but finding that raw visual features and simple textual labels often fail to capture latent intentions, as shown in Fig.~\ref{fig:motivation}(b).

Based on these insights, we propose the \textbf{Suspicion Progression Analysis Network (SPAN)}, an innovative method that models the continuous progression of hidden intentions through formal computation.
SPAN comprises two components: TPP Suspicion Modeling and Suspicion Coefficient Modulation.
TPP Suspicion Modeling formalizes a suspicion score formula that effectively models continuous changes while accounting for long-term dependencies and cumulative effects, inspired by the TPP theory. The modeling recognizes that different suspicious actions effects overall suspicion differently based on their frequency and duration, leading us to design specialized kernel functions for modeling individual suspicious action effects.
Suspicion Coefficient Modulation dedicates to modulate suspicion coefficients in the suspicion formula by leveraging multimodal information reflecting different effects of suspicious actions. Notably, we propose a Concept-Anchored Mapping method that quantifies associations between suspicious actions and predefined suspicious intention concepts, enabling understanding of not just actions occurring but also their potential underlying intentions.
By focusing on intention progression processes rather than simply classifying intention categories, SPAN significantly enhances explainability and utility. Fig.~\ref{fig:method_comparison} illustrates how our continuous SPAN differs from previous discrete method in~\cite{qi2024predicting}.

Through extensive experiments on the HAI dataset~\cite{qi2024predicting}, we demonstrate that SPAN significantly outperforms existing methods. Our method reduces MSE by 19.8\% and improves average mAP by 1.78\%. Notably, SPAN achieves a 2.74\% mAP gain in low-frequency cases, indicating that continuous models more effectively capture subtle behavioral changes.
When evaluating performance across different environments (indoor and outdoor), we observe that the advantages of continuous models are more evident in complex settings, highlighting stronger adaptability to varying scenarios.
These results support our theoretical hypothesis: the key to hidden intention localization lies in precisely modeling the dynamic progression of suspicion levels rather than applying simple classification.

Our contributions are summarized as follows:
\begin{itemize}
\item \textbf{Continuous Suspicion Modeling:} We reconceptualize hidden intention localization by introducing SPAN, a continuous method for modeling suspicion scores. This method overcomes the limitations of discrete classification and enables earlier and more accurate risk alerts by capturing the evolving nature of suspicious intentions.
\item \textbf{Point Process-Based Formulation:} We construct a formal equation based on point process theory to predict the temporal progression of suspicion scores. This formulation incorporates essential characteristics, including long-term dependencies, cumulative effects, and the frequency and duration of suspicious actions.
\item \textbf{Intention-Action Mapping:} We propose a Concept-Anchored Mapping method that associates observed actions with potential intentions. This method bridges action recognition and intention understanding, enabling context-aware reasoning similar to that of security experts and improving both system performance and explainability.
\end{itemize}

\section{Motivation}
\subsection{Suspicion Patterns for Point Processes}
\label{sec: motivation(a)}
\paragraph{Long-Term Dependencies}
To analyze the temporal progression of suspicion scores, particularly past actions' effect, we visualize the autocorrelation plot in Fig.~\ref{fig:motivation}(a) (top).
The plot illustrates temporal dependencies, with the x-axis representing lag(time difference between observations) and the y-axis showing autocorrelation values. The curve begins near 1 and gradually declines, indicating strong positive autocorrelation that weakens over time. This shows that past actions exert a lasting effect on current suspicion scores, revealing long-term dependencies.
\paragraph{Cumulative Effects}
To further examine the long-term effect of past actions, we analyze the cumulative effects of suspicion scores.
The cumulative effect curve, shown in Fig.~\ref{fig:motivation}(a) (bottom), visualizes the accumulated effects over time. The x-axis denotes the frame number and the y-axis indicates the cumulative score.
The curve increases steadily, indicating that suspicion scores accumulate over time, reflecting the sustained effect of past actions.
Thus, suspicion is shaped by a sequence of actions, leading to a progressive buildup.

\subsection{Compositional Hidden Intention} 
\label{sec: motivation(b)}
Hidden intentions are often deliberately concealed, embedding latent motivations that raw visual features may fail to capture.
To explore alternative representations of hidden intentions, we incorporate text to enrich semantic understanding.

We employ CLIP~\cite{radford2021learning} to extract textual features from suspicious action labels in the HAI dataset~\cite{qi2024predicting} and compute their similarities with visual features. As shown in Fig.~\ref{fig:motivation}(b) (right), these similarities show minimal variation, indicating that textual features do not distinguish semantic content of suspicious actions based on the visual content. Our exploration suggests that richer semantic descriptions of suspicious behavioral concepts could better differentiate visual content while more effectively recognizing hidden intentions (Fig.~\ref{fig:motivation}(b) left), motivating us to develop a method bridging observable actions and hidden motives. 

\section{Related Work}
\subsection{Temporal Action Localization}
Temporal Action Localization (TAL), also known as Temporal Action Detection, aims to accurately identify action intervals in untrimmed videos and understand human actions.  
Significant progress has been made in recent years, evolving from early proposal-based methods~\cite{escorcia2016daps,xu2017r,8237654,7780580,lin2018bsn}, which adapted two-stage object detection frameworks, to more advanced techniques enhancing proposal quality and boundary localization~\cite{lin2020fast}. More recently, innovations from object detection have influenced TAD, driving the emergence of one-stage methods~\cite{zhang2022actionformer,shi2023tridet,liu2024harnessing} and DETR-based architectures~\cite{liu2022end,zhu2024dual,9854104}.  
These pioneering TAD methods localize suspicious actions and provide their temporal information, forming the foundation for {\ourmethod}'s subsequent suspicion modeling.

\subsection{Hidden Intention}  
Traditional intention recognition research~\cite{wei2017inferring,wei2018and,liu2020spatiotemporal,fang2021learning,xu2021trajectory,xu2022gaze} focuses mainly on discovering \emph{normal intention}, referring to conscious, transparent motivations that are openly expressed and easily observable in action. This limits the ability to uncover hidden intentions before abnormal actions occur.  
Zhou \etal~\cite{zhou2023uncovering} introduced a more practical yet challenging task, Hidden Intention Discovery, which aims to recognize \emph{hidden intention}—a concealed motivation or goal preceding abnormal action. For fine-grained hidden intention recognition, Qi \etal~\cite{qi2024predicting} proposed Temporal Intention Localization (TIL), which localizes a series of pre-abnormal actions associated with hidden intentions.  
However, they measure hidden intention by quantifying its suspicion into discrete levels (uncertain, suspicious, alert), neglecting its dynamic progression. To address this gap, our method, {\ourmethod}, explores the continuous progression of hidden intention through formal computation of suspicion scores, considering the intrinsic effects of action, interactions between actions, and the effects of environmental factors.

\subsection{Temporal Point Process}
Temporal Point Process (TPP)~\cite{daley2007introduction} provides a principled framework for modeling event data and capturing interdependencies between events occurring at arbitrary temporal intervals.  
It has been widely applied in domains, including social media analysis~\cite{sharma2021identifying,zhao2015seismic}, natural language processing~\cite{zeng2023early,8624540}, and audio processing~\cite{huang2019recurrent}.  
In computer vision, Shen \etal~\cite{shen2018egocentric} applied TPP, specifically the Hawkes process~\cite{hawkes1971spectra}, to model long-range asynchronous dependencies among gaze-driven events in egocentric videos for activity prediction.  
Building upon the traditional concept of TPP, which primarily considers the timing of individual event occurrences, we propose a more comprehensive formulation by incorporating the varying effects of different events (\ie ``actions'' in our work) based on their duration and frequency.

\begin{figure*}[t]
    \centering
    \includegraphics[width=\textwidth]{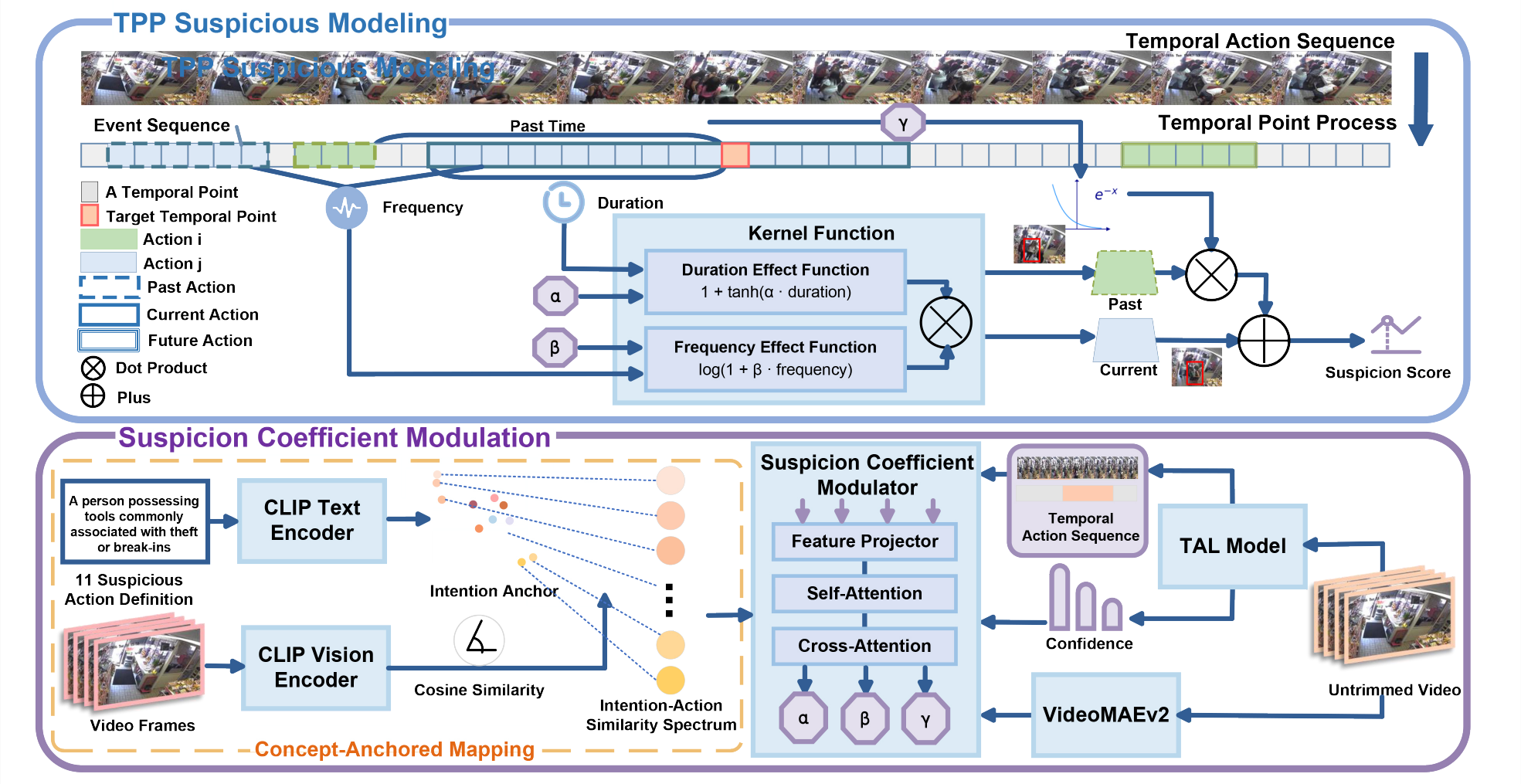}
    \caption{Overview of our Suspicion Progression Analysis Network (SPAN). The framework consists of: (1) TPP Suspicion Modeling, which formalizes suspicion scores capturing continuous changes and temporal dependencies; and (2) Suspicion Coefficient Modulation with our Concept-Anchored Mapping method that associates suspicious actions with underlying intention concepts. Together, these components effectively model the continuous progression of hidden intentions.}
    \label{fig:framework}
\end{figure*}

\section{Method} 
We propose the Suspicion Progression Analysis Network (SPAN) to model the continuous progression of hidden intentions, as illustrated in Fig.~\ref{fig:framework}. 
\subsection{Preliminary}  
\subsubsection{Temporal Action Localization}  
Temporal Action Localization (TAL) is a fundamental task in video understanding, aiming to identify the precise time intervals during which various actions occur within a video.  
In our method, we utilize a TAL model to detect 11 predefined suspicious actions. For each detected occurrence of a suspicious action, the TAL model outputs:  
\begin{itemize}  
    \item Start and end times  
    \item Confidence  
\end{itemize}
Unlike conventional TAL tasks, we not only focuses on detecting actions but also investigates the temporal progression of suspicion associated with these actions.  
Thus, the TAL model serves solely as a provider of temporal information regarding actions.  
The core challenge lies in how we analyze this temporal information and model the underlying suspicion dynamics.  
For our experiments, we adopt DyFADet~\cite{yang2024dyfadet} as the TAL model.  

\subsubsection{Temporal Point Process}  
Temporal Point Process (TPP)~\cite{daley2007introduction} is a class of stochastic processes used to model sequences of discrete events.
The core of a TPP is the conditional intensity function $\lambda(t)$, which represents the rate of occurrence of a new event at time $t$ conditioned on past events.  

The Hawkes process~\cite{hawkes1971spectra} is a specific type of point process, characterized by its self-exciting property. In the Hawkes process, the conditional intensity function is defined as:  
\begin{equation}  
\label{eq: tpp}
\lambda(t) = \mu(t) + \sum_{t_i < t}{\omega e^{-\gamma(t-t_i)}}  
\end{equation}  
where the first term represents the background intensity of the current event, independent of past events, and the second term accounts for the excitation effect of past events on the present event. Here, $t_i$ denotes the time of past event occurrences, $\omega$ is the intensity of the excitation effect, and $\gamma$ controls the decay rate.

\subsection{TPP Suspicion Modeling}
Based on the two characteristics of the suspicion score identified in Sec.~\ref{sec: motivation(a)}, we model continuous suspicion scores.
\subsubsection{Suspicion Formula}  
\label{sec: formula}
The progression of suspicion in hidden intentions follows a cumulative process driven by repeated and concurrent actions. Long-term dependencies imply that the process depends on both current and past actions, closely resembling the conditional intensity function in a Temporal Point Process (TPP).  
The Hawkes process, in particular, effectively captures the excitation relationship between past and future events, making it well-suited to model the cumulative effects of hidden intentions.
Motivated by TPP, we treat each suspicious action in the video as a discrete event, forming a temporal point process over a series of frames. We define the \textbf{\emph{suspicion score}} at time $t$ as:  
\begin{equation}  
    s(t) = \sum_{i\in\mathcal{C}_t}f(i) + \sum_{j\in\mathcal{P}_t}f(j) \cdot e^{-\gamma(t- t_j)}  
\end{equation}  
where $f(\cdot)$ is the \textbf{\emph{kernel function}} modeling the suspicion effect of an individual action. $\mathcal{C}_t$ and $\mathcal{P}_t$ denote the sets of current and past actions at time $t$, respectively. Here, $i$ and $j$ represent the $i$-th and $j$-th actions within these sets, while $\gamma$ is a decay coefficient that varies with time $t$ (discussed in Sec.~\ref{sec: modulator}). $t_j$ indicates the end time of the $j$-th past action.  
The former term measures the manifestation of hidden intentions by current actions, while the latter term models the cumulative effects of suspicion from past actions.

\subsubsection{Kernel Function}
Transcending the traditional concept of the Temporal Point Process, which considers only the timing of single event occurrences—\ie the kernel function $\mu(t)$ in Eq.~(\ref{eq: tpp}) depends solely on time $t$—we design a kernel function that accounts for the varying effects of different actions.  
Supported by Chandola \etal~\cite{chandola2009anomaly}, who describe and detect anomalous actions based on time, frequency, and context, we extract two key attributes from the TAL model outputs for kernel function design:  
\begin{itemize}  
    \item Duration $d_i$: the duration of the $i$-th action  
    \item Frequency $n_i$: the frequency of the $i$-th action  
\end{itemize}  
where $i \in \mathcal{C}$, and $\mathcal{C}$ represents the set of current or past actions.  

By incorporating these attributes, the kernel function better captures the progression of hidden intention. It is defined as:  
\begin{equation}  
    f(i) = D(i) \cdot N(i), \quad i\in \mathcal{C}  
\end{equation}  
where $D(i)$ and $N(i)$ denote the duration effect function and frequency effect function, respectively, representing the influence of an action’s duration and frequency.  
  
\paragraph{Duration Effect Function}
In the context of suspicious actions, brief actions (\eg``Quick Glance to the Side'') receive a base score, with even a slight extension significantly increasing suspicion.
For actions of moderate duration (\eg``Surveying the Area''), suspicion should rise steadily.
For prolonged actions (\eg``Staring for an Extended Period''), the rate of increase in suspicion should slow over time.  
The hyperbolic tangent function effectively captures these characteristics:  
\begin{equation}  
    D(i) = 1 + \tanh(\alpha \cdot d_i)  
\end{equation}  
where $\alpha$ is a time-dependent coefficient named duration coefficient (discussed in Sec.~\ref{sec: modulator}).  

\paragraph{Frequency Effect Function}  
The frequency effect function is designed to reflect the importance of repetitive action patterns.  
Research in security monitoring highlights that action repetitiveness is a key indicator of intention.  
However, this effect is not linear—the first occurrence of a suspicious action receives a base score, with the second and third occurrences significantly increasing suspicion, but the marginal gain from additional repetitions gradually diminishes.  
The logarithmic function precisely models this diminishing effect:  
\begin{equation}  
    N(i) = \log(1+\beta \cdot n_i)  
\end{equation}  
where $\beta$ is a time-dependent coefficient named frequency coefficient (discussed in Sec.~\ref{sec: modulator}).

\subsection{Suspicion Coefficient Modulation}
\label{sec: modulator}
In Sec.~\ref{sec: formula}, we define the formula for computing the suspicion score. However, the \textbf{\emph{decay coefficient}} $\mathbf{\gamma}$, \textbf{\emph{duration coefficient}} $\mathbf{\alpha}$, and \textbf{\emph{frequency coefficient}} $\mathbf{\beta}$, which vary with time $t$, must be dynamically modulated based on the video content.  
The optimal values of these suspicion coefficients can differ significantly across various scenarios and contexts.
To address this, we propose a \emph{\modulator} that modulates these three suspicion coefficients by integrating multimodal information extracted from video frame features, thereby enabling adaptive suspicion score prediction.

\subsubsection{Basic Multimodal Information}
\paragraph{Visual Feature}  
Hidden intention is often accompanied by specific action sequences, unusual movements, or unnatural postures.  
To predict the suspicion score, we utilize VideoMAEv2~\cite{wang2023videomae} to extract visual features $e_{visual} \in \mathbb{R}^{1408}$ as input to the modulator, thereby capturing observable actions within the video.  
Additionally, environmental information embedded in visual features aids in better recognizing actions by integrating contextual cues, ensuring that action judgments are contextually relevant.  

\paragraph{Confidence Feature}  
The confidence feature is derived from the TAL model's detection results and represented as $e_{conf} \in \mathbb{R}^{11}$, a vector corresponding to the confidence scores of the 11 suspicious actions.  This design enables the modulator to differentiate between highly certain detections and potentially misclassified detections.  

\paragraph{Temporal Feature}  
To capture fundamental temporal dynamics, we extract the point process temporal feature $e_{temp} \in \mathbb{R}^{14}$, a vector concatenated from:
\begin{itemize}  
    \item The number of currently active actions (scalar, length 1)  
    \item The number of past actions (scalar, length 1)  
    \item The distribution of action types (vector, length 11, where present actions are marked as 1 and absent actions as 0)  
    \item The timestamp feature (scalar, length 1, indicating the frame number currently being processed)  
\end{itemize}  
The temporal feature captures the patterns of action occurrences within the scene, providing essential temporal context.  

\subsubsection{Concept-Anchored Mapping Method}
As shown in Sec.~\ref{sec: motivation(b)}, hidden intention is elusive both visually and semantically, as the textual features of suspicious action labels fail to distinguish the semantic content of suspicious actions.

We predefine 11 suspicious action concepts related to hidden intention (details in \textcolor{red}{\textit{supplementary material}}) and use CLIP~\cite{radford2021learning}'s text encoder to extract textual features as \emph{intention anchors}. These anchors capture high-level semantic information, converting raw visual input into meaningful concepts (\eg ``a person carefully looking around to observe the environment'' instead of merely ``looking around''). By providing contextual cues and causal interpretations, intention anchors bridge observable actions and hidden intention.

We then compute similarities between CLIP's visual features and these intention anchors, mapping the video's relevance to these anchors onto an \emph{intention-action similarity spectrum}, denoted as $e_{seman} \in \mathbb{R}^{11}$.  
This spectrum provides an interpretable representation, indicating which suspicious concepts an action frame resembles and to what extent, thereby enhancing the transparency of the model’s decision-making.  
In addition, it effectively bridges visual observations and semantic understanding, enabling the system to ``comprehend'' action rather than merely ``observe'' it.
The spectrum is also input to the {\modulator}.

\subsubsection{\modulator}
Our {\modulator} consists of three main components: a feature projector, a multimodal fusion module, and a coefficient-specific cross-attention mechanism.

\paragraph{Feature Projector}  
We design dedicated feature projectors for each input, projecting them into a unified representation space through nonlinear transformations. These feature projectors not only standardize the dimensions of multimodal features but also enhance their expressiveness through nonlinear transformations.

\paragraph{Multimodal Fusion Module}  
After feature projection, we employ a simplified attention mechanism to adaptively integrate information from different modalities. The attention weight for each modality is computed using a learnable projection matrix, followed by weighted aggregation and fusion through a fully connected layer to produce the final fused feature $h$. This mechanism allows the model to dynamically focus on the most relevant modalities in different scenarios, such as prioritizing visual features in scenes with prominent visual cues or temporal features when action patterns are more informative.

\paragraph{Coefficient-Specific Cross-Attention Mechanism}  
To ensure the three temporal suspicion coefficients focus on distinct aspects of fused features, we introduce a coefficient-specific cross-attention mechanism. For each coefficient $p \in \{\gamma, \alpha, \beta\}$, we initialize a coefficient-specific query vector $q_p$ that processes alongside the fused feature $h$ via cross-attention, yielding a \textbf{context vector} for each coefficient: 
\begin{equation}
    c_p = \text{softmax}\left(\frac{Q_{p}K^T}{\sqrt{d_k}}\right)V
\end{equation}
where $Q_p$ is computed as the dot product of $q_p$ and a learnable weight matrix $W_p$, while $K$ and $V$ are derived from the fused feature $h$.

The \textbf{modulation factor} for each coefficient is generated as:  
\begin{equation}
    \delta_p = \tanh{(W_p^T c_p + b_p)} \cdot 0.5
\end{equation}
The modulation factor is constrained within the range $[-0.5, 0.5]$.  

Finally, the dynamically modulated \textbf{coefficient} is computed as:  
\begin{equation}
    p = \omega_p \cdot (1+\delta_p)
\end{equation}
where $\omega_p$ is the base value for each coefficient.  
This formula ensures that coefficient values remain within a reasonable range while allowing for dynamic modulation based on varying scenarios and contexts. For instance, in high-security environments, the system may increase the base value $\omega_{\beta}$ to heighten sensitivity to transient actions and raise $\omega_{\gamma}$ to extend the influence of past actions.  
Conversely, in low-risk areas, the system may reduce these base values to lower the false alarm rate.  

\subsubsection{Loss Function}
We propose a \textbf{\emph{wave-aware loss}} consisting of three components: base loss $\mathcal{L}_{base}$ (Smooth L1 Loss for value disparity), magnitude loss $\mathcal{L}_{magn}$ (MSE of first-order differences for amplitude changes), and trend loss $\mathcal{L}_{trend}$ (direction matching for pattern preservation):
\begin{equation}
\mathcal{L} = \mathcal{L}_{base} + \lambda_{magn}\mathcal{L}_{magn} + \lambda_{trend}\mathcal{L}_{trend}
\end{equation}
with optimal weights $\lambda_{magn}=0.5$ and $\lambda_{trend}=0.3$. This multi-objective function balances point-level accuracy and sequence-level dynamic pattern modeling. Detailed analysis is provided in the \textcolor{red}{\textit{supplementary material}}.

\begin{table*}[t]
\centering
\caption{Performance comparison of our method against the baseline on the HAI dataset. The best scores in each column are highlighted in bold. Results of our method are shaded in gray.}
\label{tab:main_results}
\begin{tabular}{l|ccc|cccccc}
\hline
\multirow{2}{*}{Method} & \multicolumn{3}{c|}{Curve Fitting} & \multicolumn{6}{c}{Temporal Localization (mAP$\uparrow$ [\%])} \\
 & MSE$\downarrow$ & MAE$\downarrow$ & R$^2\uparrow$ & IoU=0.3 & IoU=0.4 & IoU=0.5 & IoU=0.6 & IoU=0.7 & Avg. \\
\hline
Baseline & 0.206 & 0.442 & -16.761 & 9.82 & 7.12 & 4.96 & 2.45 & 1.42 & 5.15 \\
\cellcolor{gray!20}\textbf{Ours} & \cellcolor{gray!20}\textbf{0.008} & \cellcolor{gray!20}\textbf{0.059} & \cellcolor{gray!20}\textbf{0.100} & \cellcolor{gray!20}\textbf{12.08} & \cellcolor{gray!20}\textbf{9.31} & \cellcolor{gray!20}\textbf{6.72} & \cellcolor{gray!20}\textbf{4.10} & \cellcolor{gray!20}\textbf{2.43} & \cellcolor{gray!20}\textbf{6.93} \\
\hline
\end{tabular}
\end{table*}

\begin{figure*}[t]
    \centering
    \label{tab:scenarios}{
        \begin{tabular}{l|l|ccc|cccccc}
        \hline
        \multirow{2}{*}{Scenario} & \multirow{2}{*}{Method} & \multicolumn{3}{c|}{Curve Fitting} & \multicolumn{6}{c}{Temporal Localization (mAP$\uparrow$ [\%])} \\
         & & MSE$\downarrow$ & MAE$\downarrow$ & R$^2\uparrow$ & IoU=0.3 & IoU=0.4 & IoU=0.5 & IoU=0.6 & IoU=0.7 & Avg. \\
        \hline
        \multirow{2}{*}{Indoor} & Baseline & 0.218 & 0.482 & -17.843 & 9.04 & 6.63 & 4.45 & 2.11 & 1.25 & 4.70 \\
         & \cellcolor{gray!20}\textbf{Ours} & \cellcolor{gray!20}\textbf{0.008} & \cellcolor{gray!20}\textbf{0.063} & \cellcolor{gray!20}\textbf{0.097} & \cellcolor{gray!20}\textbf{11.78} & \cellcolor{gray!20}\textbf{8.98} & \cellcolor{gray!20}\textbf{6.07} & \cellcolor{gray!20}\textbf{3.87} & \cellcolor{gray!20}\textbf{1.95} & \cellcolor{gray!20}\textbf{6.53} \\
        \hline
        \multirow{2}{*}{Outdoor} & Baseline & 0.133 & 0.299 & -9.529 & 10.53 & 8.15 & 5.87 & 3.12 & 1.79 & 5.61 \\
         & \cellcolor{gray!20}\textbf{Ours} & \cellcolor{gray!20}\textbf{0.007} & \cellcolor{gray!20}\textbf{0.055} & \cellcolor{gray!20}\textbf{0.133} & \cellcolor{gray!20}\textbf{13.46} & \cellcolor{gray!20}\textbf{10.30} & \cellcolor{gray!20}\textbf{6.78} & \cellcolor{gray!20}\textbf{4.51} & \cellcolor{gray!20}\textbf{3.57} & \cellcolor{gray!20}\textbf{7.72} \\
        \hline
        \multirow{2}{*}{$<$50\% Frequency} & Baseline & 0.119 & 0.235 & -8.742 & 13.60 & 9.97 & 7.02 & 5.05 & 2.09 & 7.60 \\
         & \cellcolor{gray!20}\textbf{Ours} & \cellcolor{gray!20}\textbf{0.006} & \cellcolor{gray!20}\textbf{0.042} & \cellcolor{gray!20}\textbf{0.163} & \cellcolor{gray!20}\textbf{16.75} & \cellcolor{gray!20}\textbf{13.63} & \cellcolor{gray!20}\textbf{11.20} & \cellcolor{gray!20}\textbf{7.41} & \cellcolor{gray!20}\textbf{2.71} & \cellcolor{gray!20}\textbf{10.34} \\
        \hline
        \multirow{2}{*}{$>$50\% Frequency} & Baseline & 0.093 & 0.223 & -7.477 & 12.78 & 10.57 & 7.23 & 4.39 & 1.73 & 7.34 \\
         & \cellcolor{gray!20}\textbf{Ours} & \cellcolor{gray!20}\textbf{0.005} & \cellcolor{gray!20}\textbf{0.064} & \cellcolor{gray!20}\textbf{0.157} & \cellcolor{gray!20}\textbf{15.95} & \cellcolor{gray!20}\textbf{10.68} & \cellcolor{gray!20}\textbf{8.93} & \cellcolor{gray!20}\textbf{7.13} & \cellcolor{gray!20}\textbf{4.95} & \cellcolor{gray!20}\textbf{9.53} \\
        \hline
        \end{tabular}
    }
    \label{fig:metrics_visualization}{
        \includegraphics[width=0.97\textwidth]{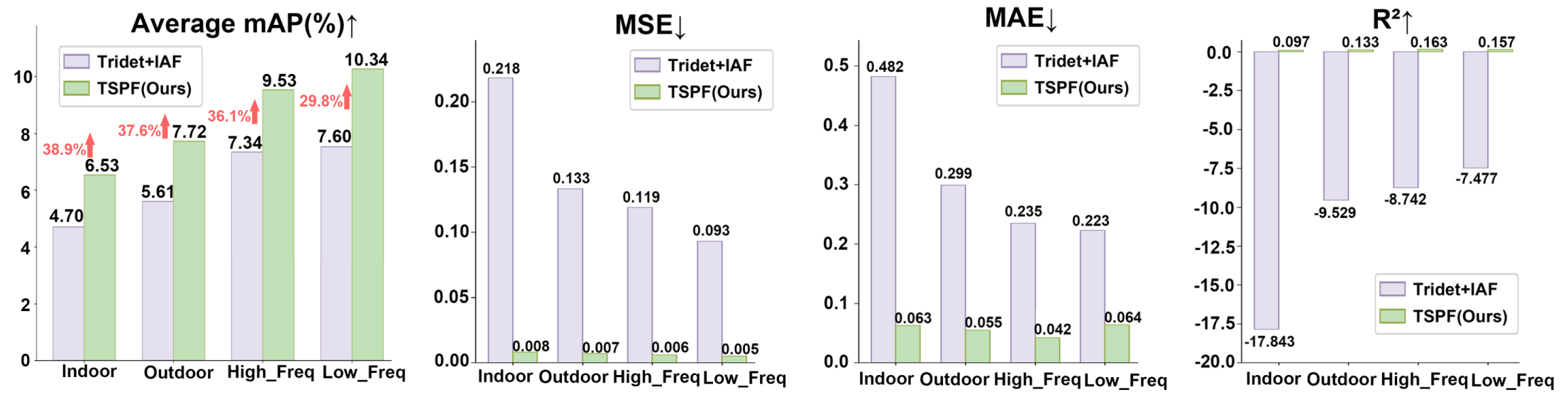}
    }
    \caption{Performance comparison on different HAI scenario subsets (table) and their visualization (graph). The best scores in each column are highlighted in bold. Results of our method are shaded in gray.}
    \label{fig:combined}
\end{figure*}

\begin{table*}[t]
  \suppressfloats[t]
\centering
\caption{Comparison between using predefined suspicious concepts (Definition) and action labels (Label). The best scores in each column are highlighted in bold. Results of our method are shaded in gray.}
\label{tab:concept_comparison}
\begin{tabular}{c|ccc|cccccc}
\hline
\multirow{2}{*}{Feature Type} & \multicolumn{3}{c|}{Curve Fitting} & \multicolumn{6}{c}{Temporal Localization (mAP$\uparrow$ [\%])} \\
& MSE$\downarrow$ & MAE$\downarrow$ & R$^2\uparrow$ & IoU=0.3 & IoU=0.4 & IoU=0.5 & IoU=0.6 & IoU=0.7 & Avg. \\
\hline
Label & 0.011 & 0.081 & 0.012 & 9.88 & 7.37 & 5.81 & \textbf{4.43} & \textbf{2.63} & 6.02 \\
Definition & \cellcolor{gray!20}\textbf{0.008} & \cellcolor{gray!20}\textbf{0.059} & \cellcolor{gray!20}\textbf{0.100} & \cellcolor{gray!20}\textbf{12.08} & \cellcolor{gray!20}\textbf{9.31} & \cellcolor{gray!20}\textbf{6.72} & \cellcolor{gray!20}4.10 & \cellcolor{gray!20}2.43 & \cellcolor{gray!20}\textbf{6.93} \\
\hline
\end{tabular}
\end{table*}

\section{Experiments}
We conduct extensive experiments to investigate the following research questions:

1) \textbf{RQ1:} Can our method effectively model suspicious intention progression and achieve good performance in temporal intention localization, despite their hidden nature?

2) \textbf{RQ2:} How robust is the performance of our method across different scenarios (indoor/outdoor environments and high/low-frequency suspicious actions)?

3) \textbf{RQ3:} What are the effects of different semantic feature extraction methods?

4) \textbf{RQ4:} In the multimodal modulator, are different modalities and their various combinations effective?

5) \textbf{RQ5:} How do different loss function components and their weights impact the model's ability to predict suspicion score curves?
\subsection{Experimental Setup}
\subsubsection{Datasets}
We conduct experiments on the challenging HAI dataset~\cite{qi2024predicting}, the established benchmark for Temporal Intention Localization. This dataset contains comprehensive annotations specifically designed for intention analysis, including 11 suspicious action categories, 3 levels of suspicious intention (uncertain, suspicious, alert), and frame-by-frame suspicion score annotations. The dataset consists of 205 untrimmed videos (1,689 action instances) for training and 23 untrimmed videos (242 action instances) for testing.

To evaluate our method's generalization capability across real-world scenarios, we further experiment on four HAI subsets: \begin{itemize}
    \item \textbf{HAI-indoors}: 115 training videos (974 action instances) and 12 testing videos (92 action instances) captured in controlled indoor environments.
    \item \textbf{HAI-outdoors}: 70 training videos (581 action instances) and 7 testing videos (46 action instances) captured in more complex outdoor environments.
    \item \textbf{HAI-high\_frequency}: 105 training videos (935 action instances) and 12 testing videos (91 action instances), where suspicious actions above 50\% of duration, representing scenarios with persistent suspicious actions.
    \item \textbf{HAI-low\_frequency}: 99 training videos (814 action instances) and 12 testing videos (91 action instances), where suspicious actions below 50\% of duration, representing challenging scenarios with sparse suspicious actions.
\end{itemize}

\subsubsection{Evaluation Metrics}
We use a dual evaluation strategy to assess both regression accuracy and classification performance:

\paragraph{Curve Fitting Metrics}
To evaluate how precisely our method captures the continuous progression of suspicion scores, we use Mean Absolute Error (MAE), Mean Squared Error (MSE), and Coefficient of Determination (R$^2$).

\paragraph{Temporal Localization Metrics}
Following the established protocol in \citeauthor{qi2024predicting}~\cite{qi2024predicting}, we categorize suspicion scores into three distinct levels (uncertain: 0-0.3, suspicious: 0.3-0.6, alert: $>$0.6) and calculate mean Average Precision (mAP) at IoU thresholds of [0.3, 0.4, 0.5, 0.6, 0.7]. This enables direct comparison with existing discrete classification methods while demonstrating the advantages of our continuous modeling method.

All evaluations are performed on the test set using VideoMAEv2 as the feature extractor for fair comparison.

\subsection{Main Results (RQ1, RQ2 \& RQ3)}
\subsubsection{Performance on HAI Dataset}
Table~\ref{tab:main_results} presents a comparison of our method against TriDet+IAF~\cite{qi2024predicting}, which represents the current state-of-the-art method for the TIL task. Our method consistently outperforms this leading baseline across all evaluation categories. In curve fitting metrics, our method demonstrates substantial improvement with a remarkable 19.8\% reduction in MSE and transforms the R$^2$ value from negative to positive, indicating a fundamentally superior modeling capability. For temporal localization performance, our method achieves consistently higher mAP scores across all IoU thresholds, yielding an overall average mAP improvement of 1.78\% over the previously best-performing method in this challenging task domain.

\subsubsection{Indoor vs. Outdoor Performance Comparison}

We conducted additional experiments on the HAI-indoors and HAI-outdoors subsets, as shown in Fig~\ref{fig:combined}. Our method outperforms the baseline in both indoor and outdoor environments, showing improvements across all metrics. Notably, the method performs better in outdoor environments, achieving a 2.21\% higher average mAP compared to 1.83\% in indoor scenarios, likely due to the richer contextual information and more diverse action patterns present in these scenes, which benefit from the multimodal fusion mechanism.

\subsubsection{High-Frequency vs. Low-Frequency Performance Comparison}

Experiments on the HAI-high\_frequency and HAI-low\_frequency subsets, detailed in Fig~\ref{fig:combined}, further show the robustness of our method. In both cases, our method yields an average mAP increase of 2.74\% for low-frequency and 2.19\% for high-frequency scenarios. Notably, the method performs even better in low-frequency cases, suggesting that our temporal point process model effectively captures implicit correlations in sparse events, which is crucial for detecting high-risk, low-frequency suspicious actions.

\subsubsection{Impact of Semantic Feature Extraction Methods}

To investigate the effectiveness of our proposed Concept-Anchored Mapping Method for semantic feature extraction, we conducted a comparative experiment between using predefined suspicious concepts (Definition) and directly using action labels (Label). Table~\ref{tab:concept_comparison} presents the performance comparison across various metrics. The results show that the definition-based method outperforms the label-based method with a 0. 91\% higher average mAP and substantially better curve fitting metrics. These findings confirm that predefined suspicious concepts provide richer semantic representations that enable the model to better comprehend suspicious actions and their underlying intentions, rather than merely observing visual patterns.

\begin{figure}[t]
    \centering
    \begin{subfigure}[b]{0.15\textwidth}
        \centering
        \includegraphics[width=\textwidth]{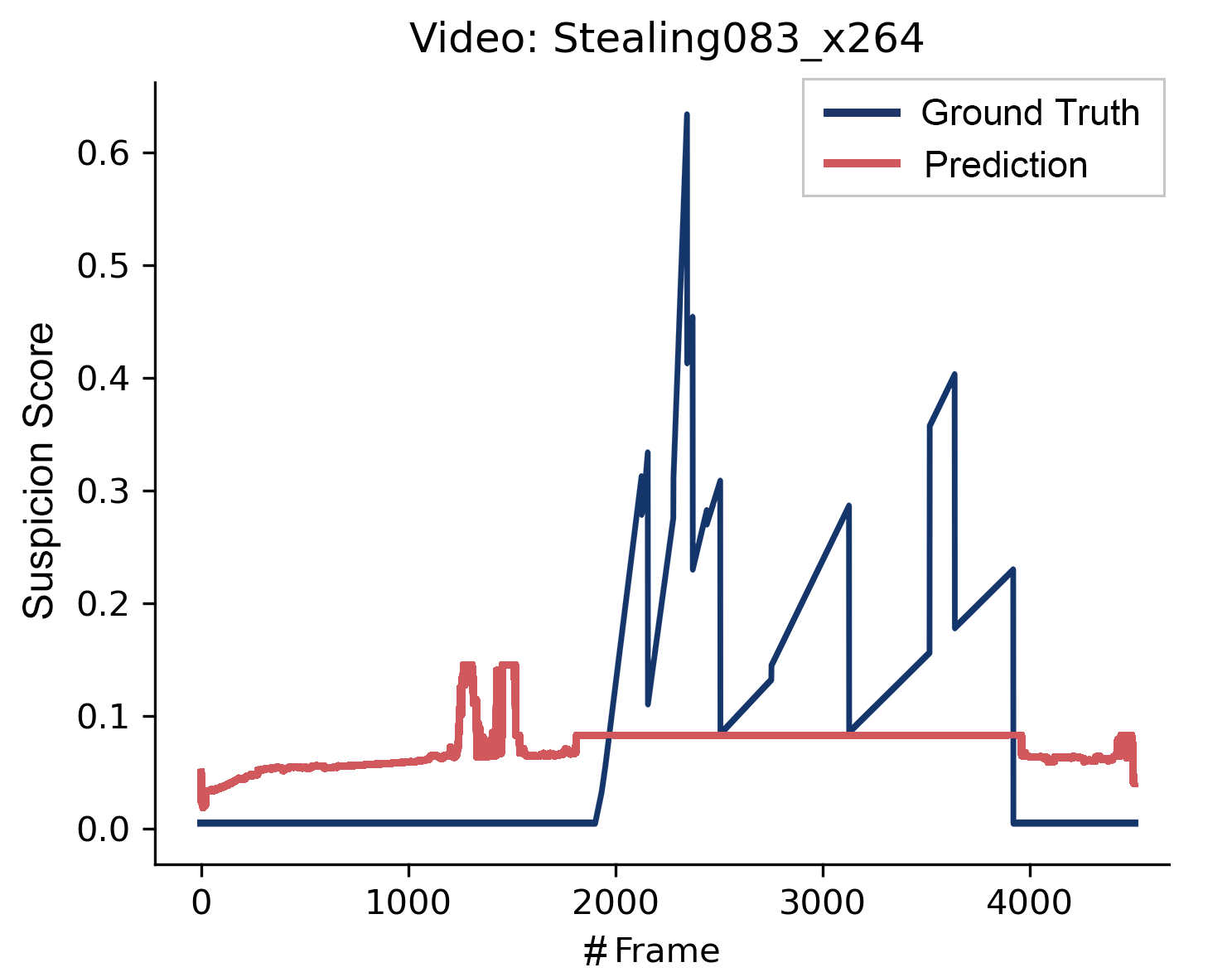}
        \caption{Baseline}
    \end{subfigure}
    \hfill
    \begin{subfigure}[b]{0.15\textwidth}
        \centering
        \includegraphics[width=\textwidth]{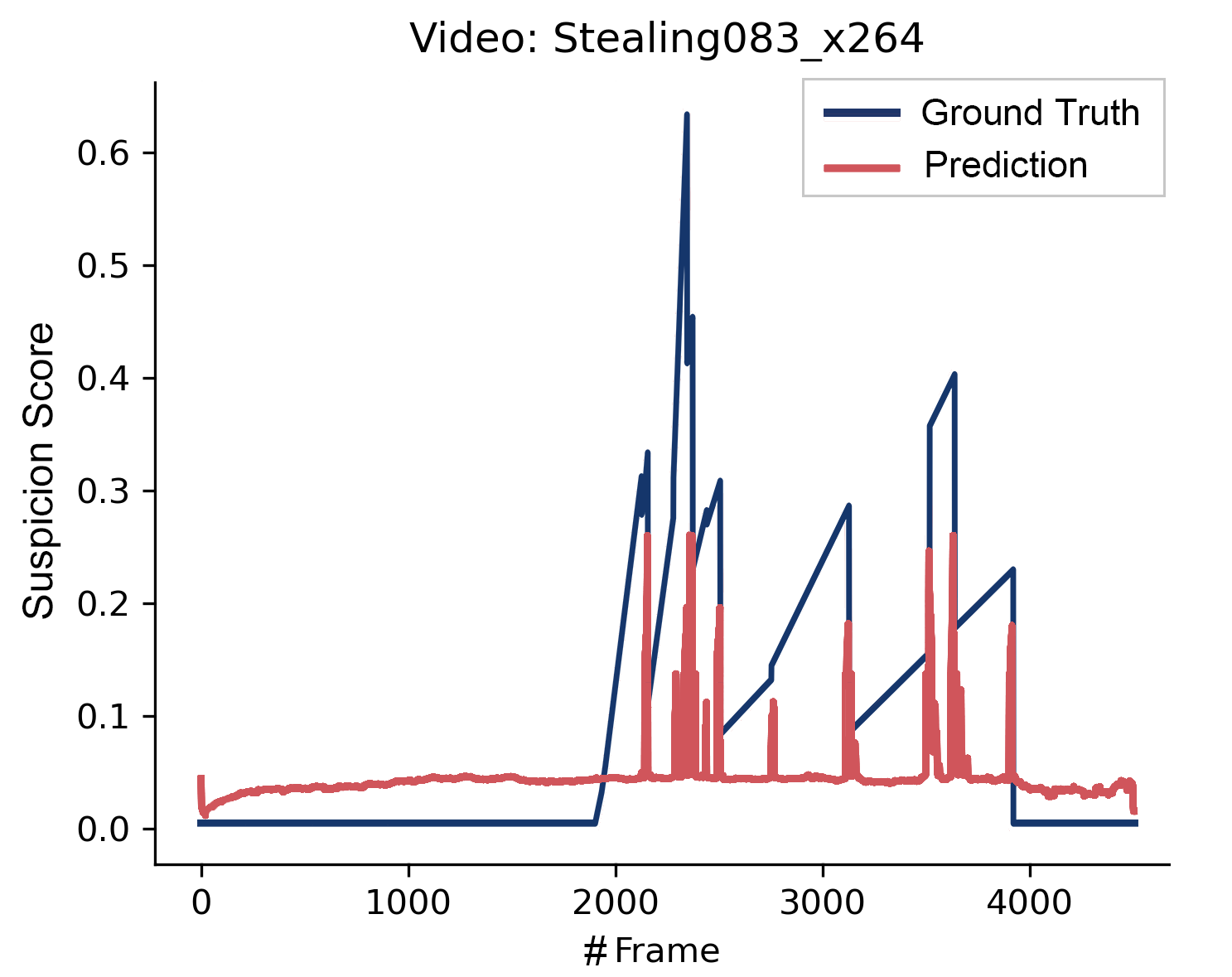}
        \caption{No mod.}
    \end{subfigure}
    \hfill
    \begin{subfigure}[b]{0.15\textwidth}
        \centering
        \includegraphics[width=\textwidth]{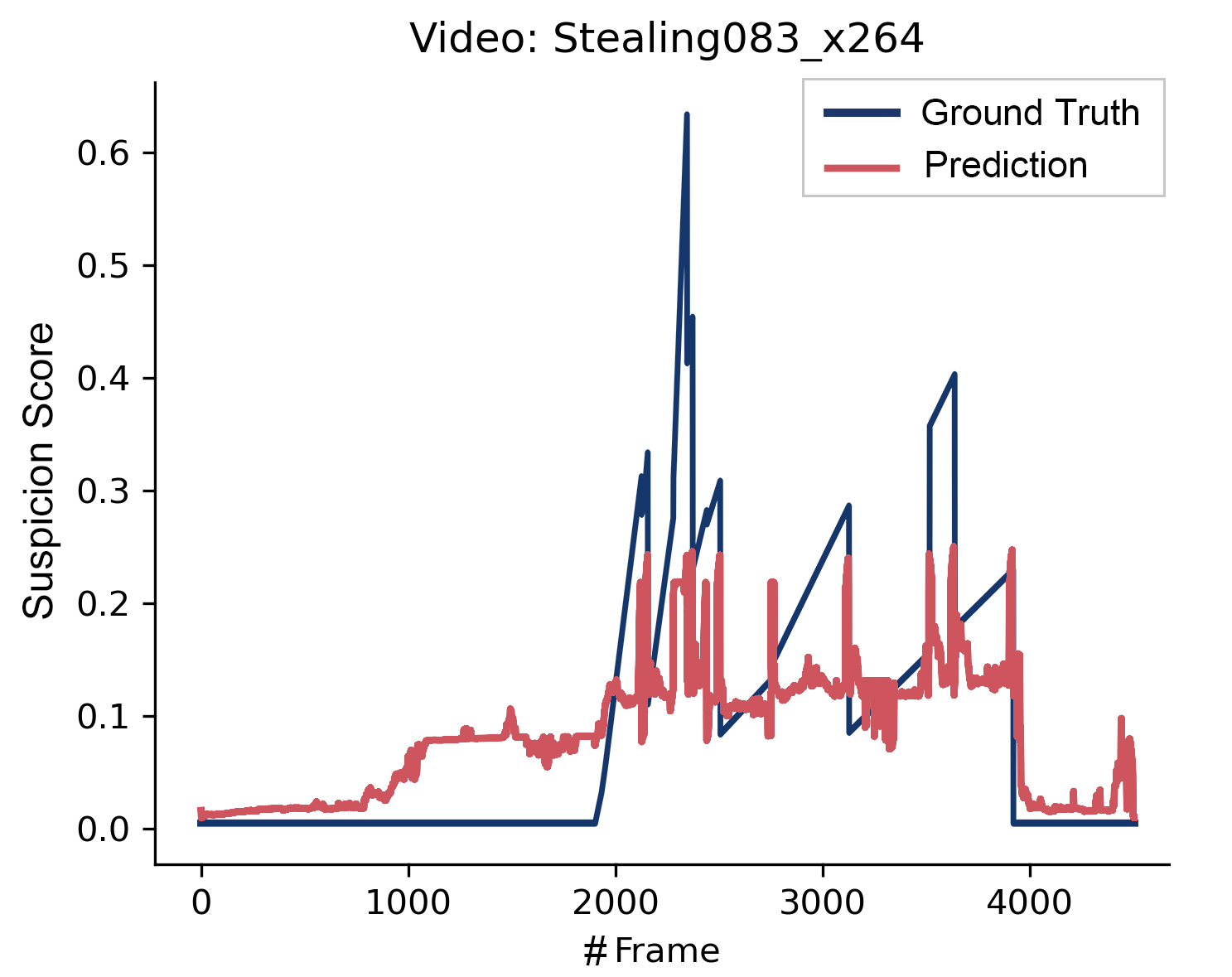}
        \caption{Conf.}
    \end{subfigure}
    
    \vspace{0.3em}
    
    \begin{subfigure}[b]{0.15\textwidth}
        \centering
        \includegraphics[width=\textwidth]{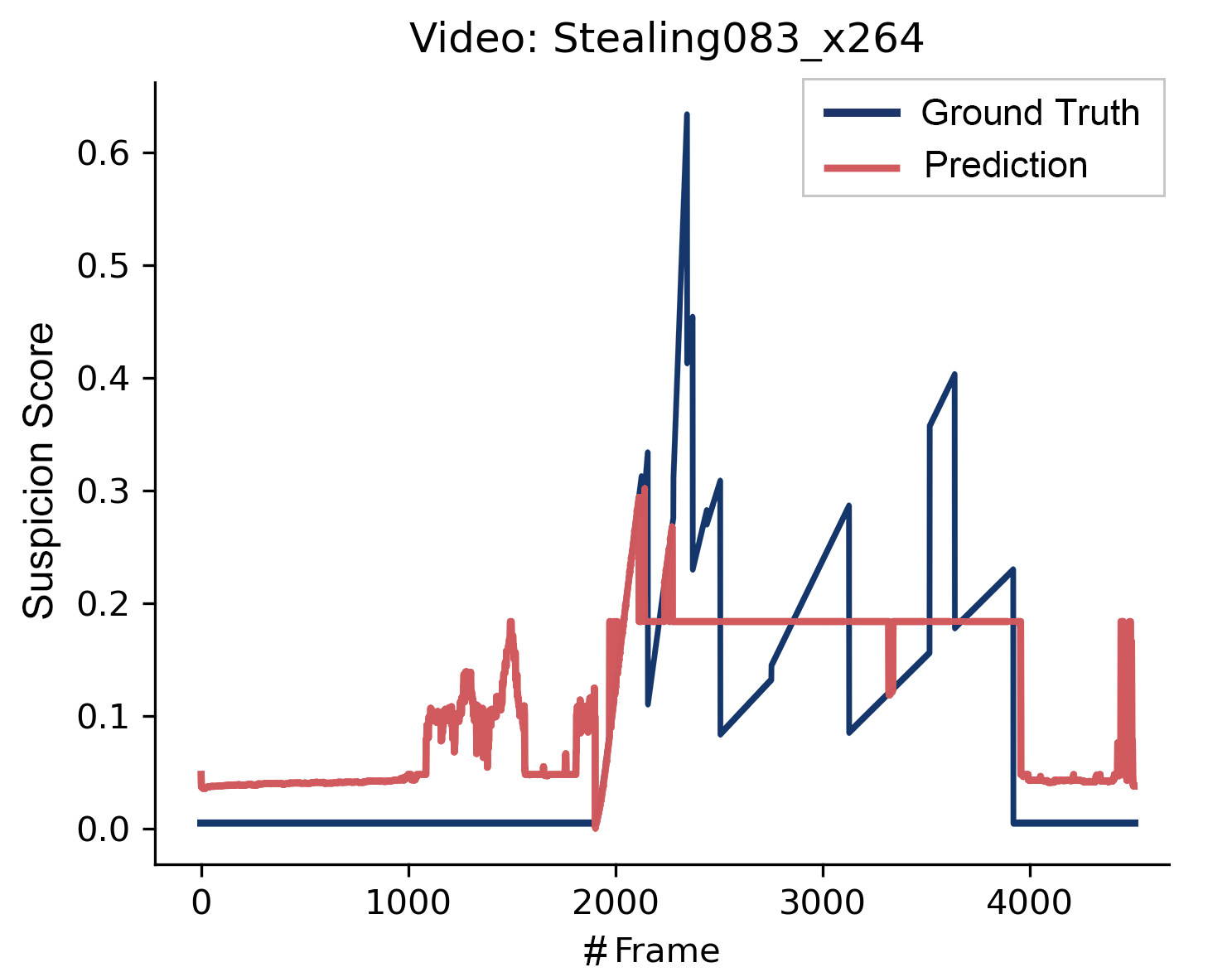}
        \caption{Semantic}
    \end{subfigure}
    \hfill
    \begin{subfigure}[b]{0.15\textwidth}
        \centering
        \includegraphics[width=\textwidth]{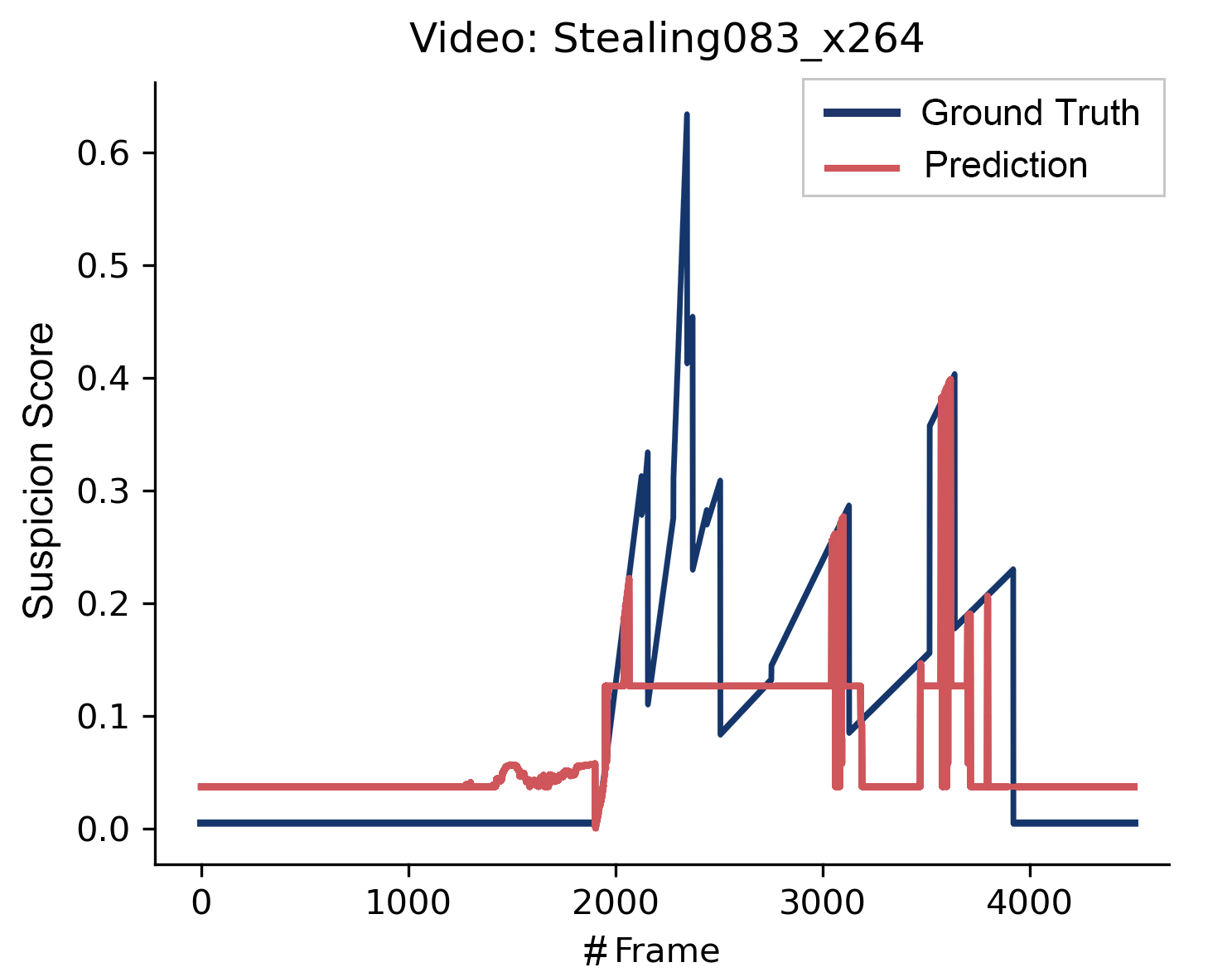}
        \caption{Visual}
    \end{subfigure}
    \hfill
    \begin{subfigure}[b]{0.15\textwidth}
        \centering
        \includegraphics[width=\textwidth]{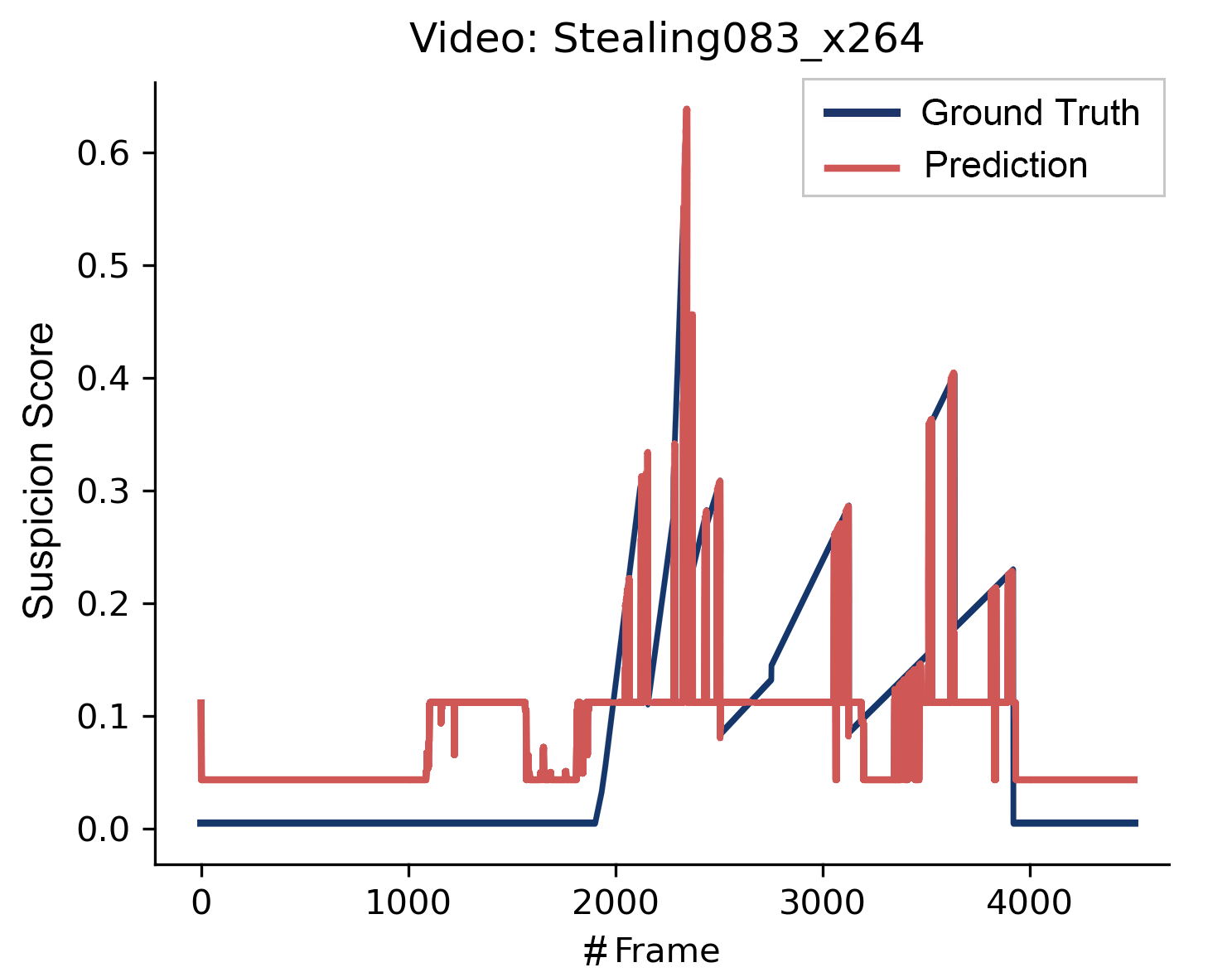}
        \caption{C+S}
    \end{subfigure}   
    \vspace{0.3em}
    \begin{subfigure}[b]{0.15\textwidth}
        \centering
        \includegraphics[width=\textwidth]{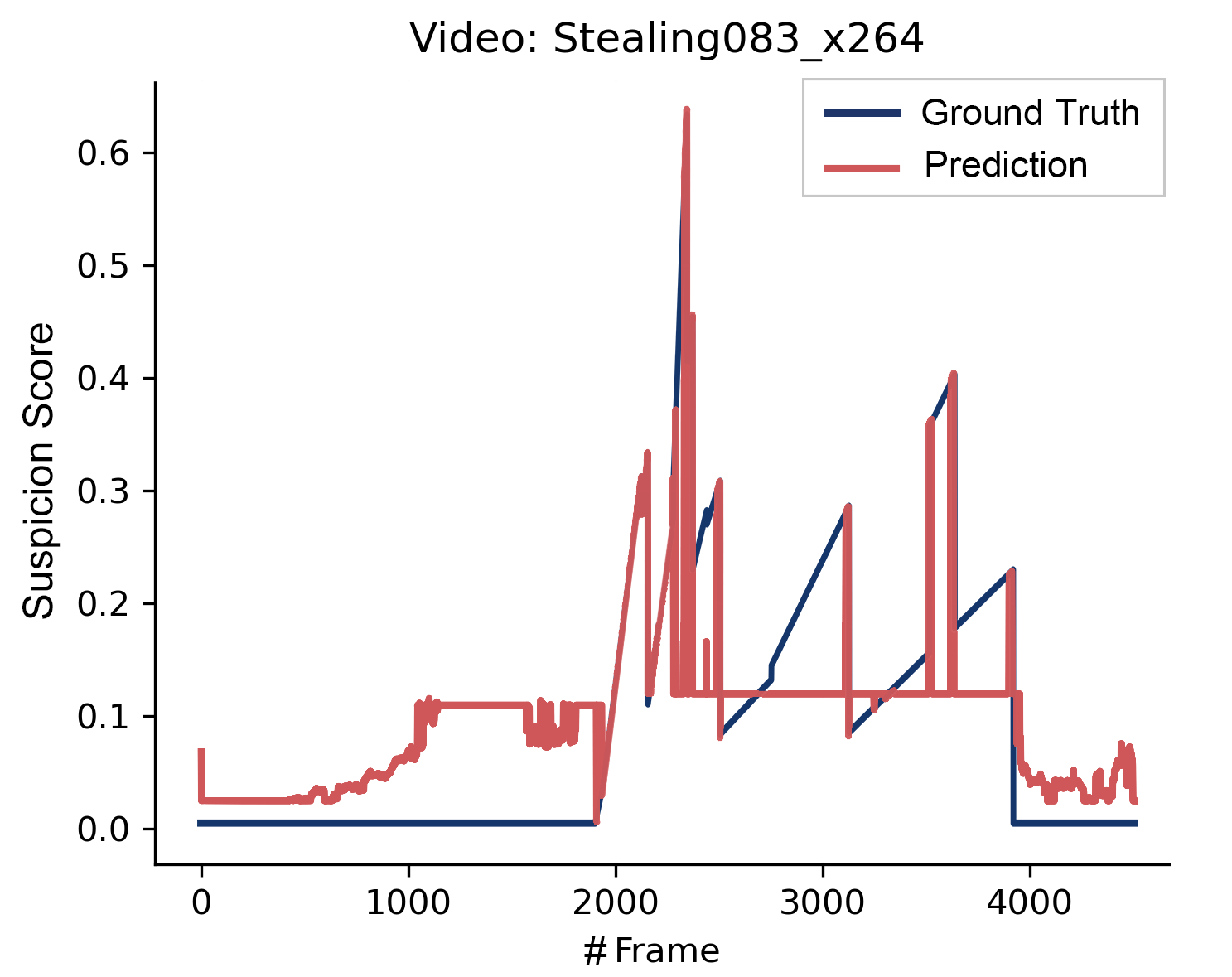}
        \caption{C+V}
    \end{subfigure}
    \hfill
    \begin{subfigure}[b]{0.15\textwidth}
        \centering
        \includegraphics[width=\textwidth]{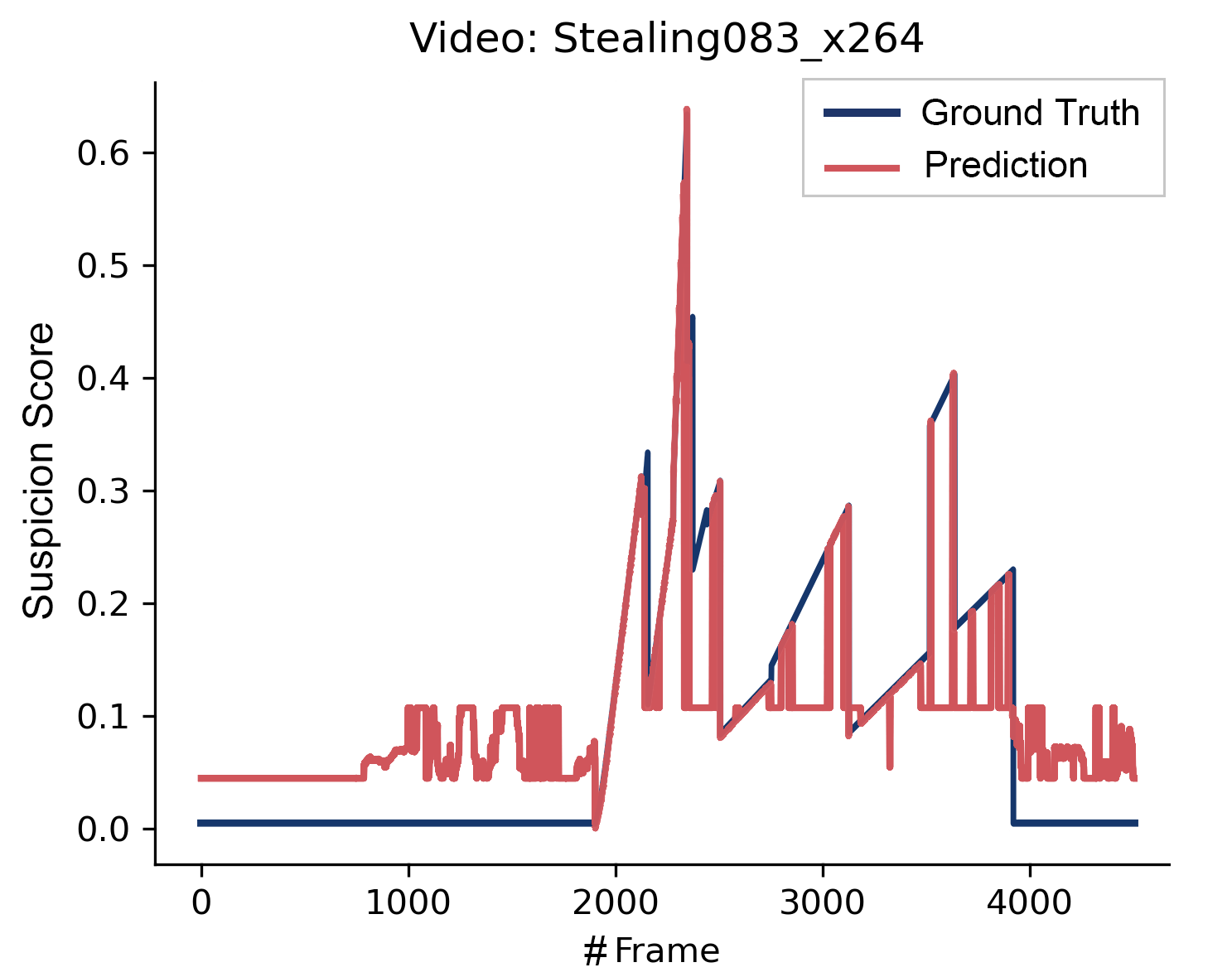}
        \caption{S+V}
    \end{subfigure}
    \hfill
    \begin{subfigure}[b]{0.15\textwidth}
        \centering
        \includegraphics[width=\textwidth]{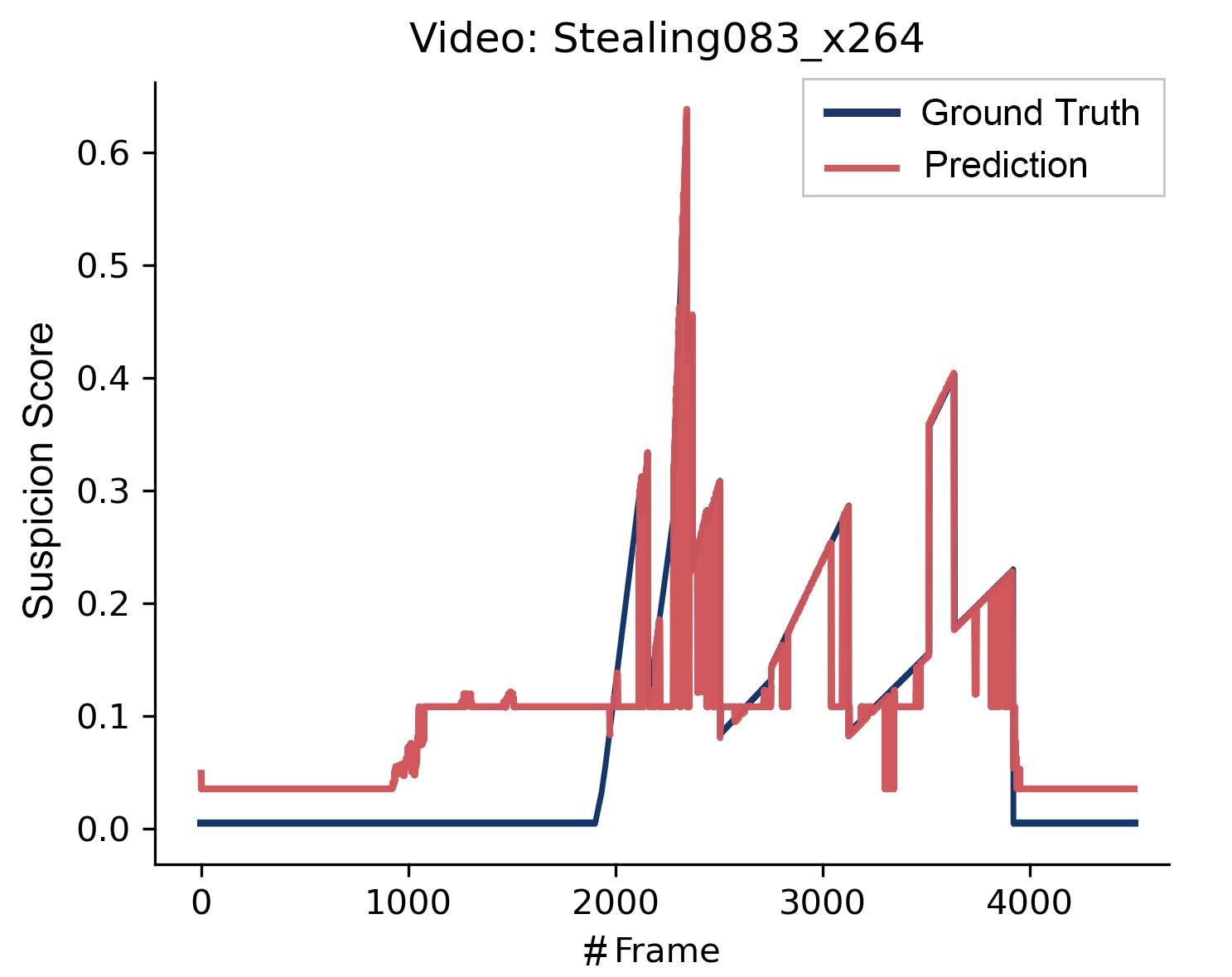}
        \caption{All}
    \end{subfigure}
    \caption{Visualization of suspicion scores with baseline and different modulation methods.}
    \label{fig:modulation_visualization}
\end{figure}
\begin{table*}[t]
  \suppressfloats[t]
\centering
\caption{Ablation study on different modal features. The best scores in each column are highlighted in bold. Results of our method are shaded in gray.}
\label{tab:feature_ablation}
\begin{tabular}{l|ccc|cccccc}
\hline
\multirow{2}{*}{Features} & \multicolumn{3}{c|}{Curve Fitting} & \multicolumn{6}{c}{Temporal Localization (mAP$\uparrow$ [\%])} \\
 & MSE$\downarrow$ & MAE$\downarrow$ & R$^2\uparrow$ & IoU=0.3 & IoU=0.4 & IoU=0.5 & IoU=0.6 & IoU=0.7 & Avg. \\
\hline
Baseline & 0.206 & 0.442 & -16.761 & 9.82 & 7.12 & 4.96 & 2.45 & 1.42 & 5.15 \\
No coefficient modulation & 0.215 & 0.432 & -15.961 & 9.79 & 6.22 & 5.06 & 2.81 & 1.76 & 5.13 \\
Confidence ($e_{conf}$) & 0.012 & 0.082 & -0.001 & 10.23 & 7.86 & 5.42 & 2.93 & 1.64 & 5.62 \\
Spectrum ($e_{seman}$) & 0.236 & 0.481 & -18.762 & 6.79 & 5.82 & 3.46 & 1.03 & 0.42 & 3.50 \\
Visual ($e_{visual}$) & 0.072 & 0.196 & -5.012 & 9.31 & 6.94 & 4.77 & 2.32 & 1.23 & 4.91 \\
$e_{conf}$ + $e_{seman}$ & 0.011 & 0.067 & 0.006 & 11.02 & 8.43 & 5.92 & 3.36 & 1.85 & 6.12 \\
$e_{conf}$ + $e_{visual}$ & 0.009 & 0.063 & 0.084 & 11.58 & 8.89 & 6.25 & 3.64 & 2.09 & 6.49 \\
$e_{seman}$ + $e_{visual}$ & 0.070 & 0.185 & -4.080 & 9.78 & 7.37 & 4.58 & 2.69 & 1.56 & 5.21 \\
\cellcolor{gray!20}\textbf{$e_{conf}$ + $e_{seman}$ + $e_{visual}$} & \cellcolor{gray!20}\textbf{0.008} & \cellcolor{gray!20}\textbf{0.059} & \cellcolor{gray!20}\textbf{0.100} & \cellcolor{gray!20}\textbf{12.08} & \cellcolor{gray!20}\textbf{9.31} & \cellcolor{gray!20}\textbf{6.72} & \cellcolor{gray!20}\textbf{4.10} & \cellcolor{gray!20}\textbf{2.43} & \cellcolor{gray!20}\textbf{6.93} \\
\hline
\end{tabular}
\end{table*}

\subsection{Ablation Studies(RQ4 \& RQ5)}
\subsubsection{Multimodal Feature Analysis}
In order to assess the contributions of various modal features to model performance, we conducted ablation experiments as presented in Table~\ref{tab:feature_ablation}. The results show that confidence features yield the best performance, followed by visual features, while semantic features alone perform poorly. Combining confidence and visual features shows impressive results, with the combination of all three features achieving optimal performance. These findings demonstrate the complementary nature of multimodal information in improving suspicion score prediction accuracy. As shown in Fig.~\ref{fig:modulation_visualization}, taking Stealing083\_x264 as an example, we present the visualization curves of both the baseline and different modality modulation methods.

\subsubsection{Loss Function Analysis}
We examine the impact of various loss function combinations on model performance, with detailed ablation studies presented in the \textcolor{red}{\textit{supplementary material}}. Our experiments confirm that the full combination of all three loss components achieves the best performance, with optimal weights of $\lambda_{magn}=0.5$ and $\lambda_{trend}=0.3$.

\begin{figure}[t]
   \centering
   \begin{subfigure}{0.48\linewidth}
       \centering
       \includegraphics[width=\textwidth]{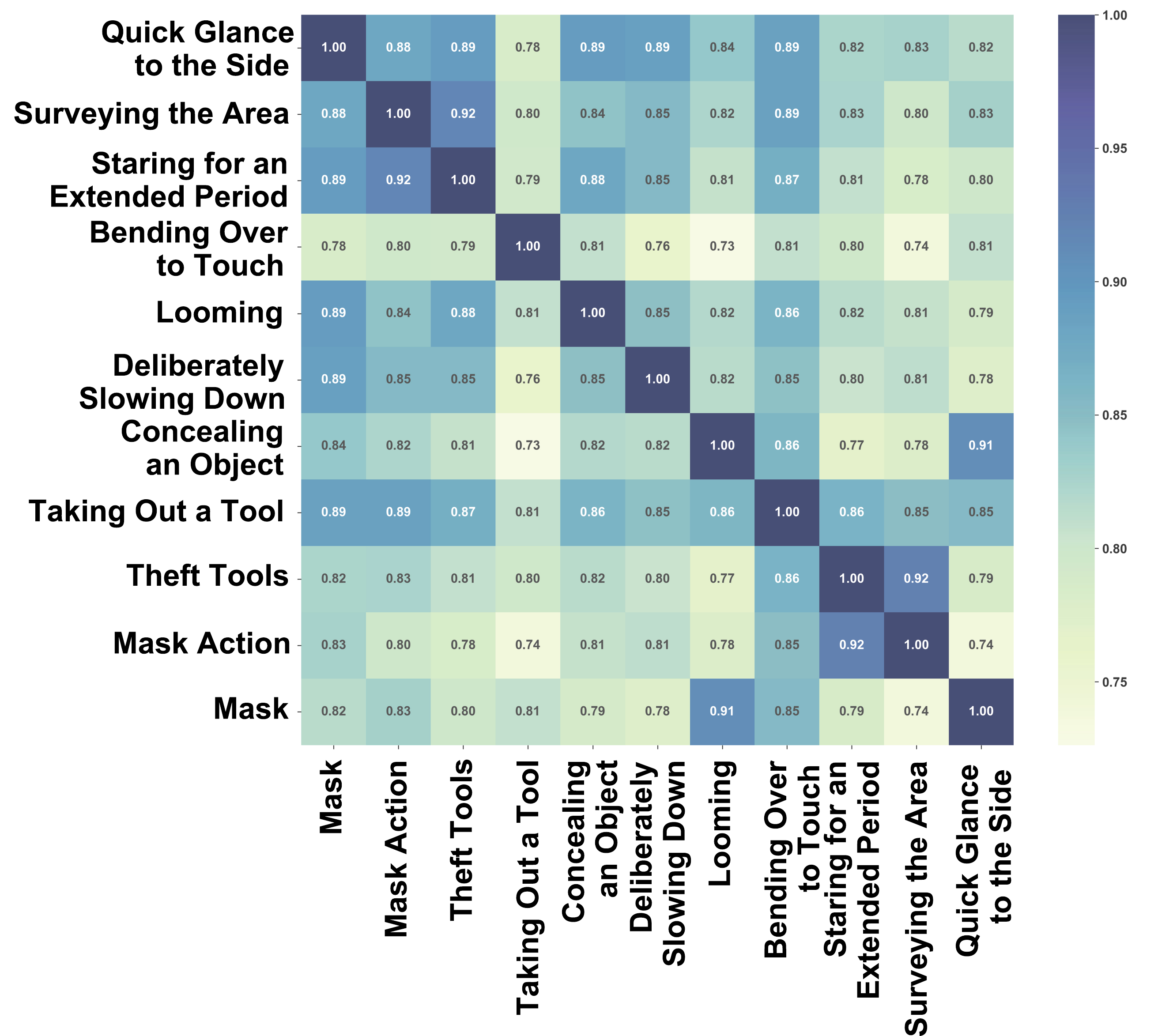}
       \caption{}
       \label{fig:similarity_heatmap}
   \end{subfigure}
   \hfill
   \begin{subfigure}{0.48\linewidth}
       \centering
       \includegraphics[width=\textwidth]{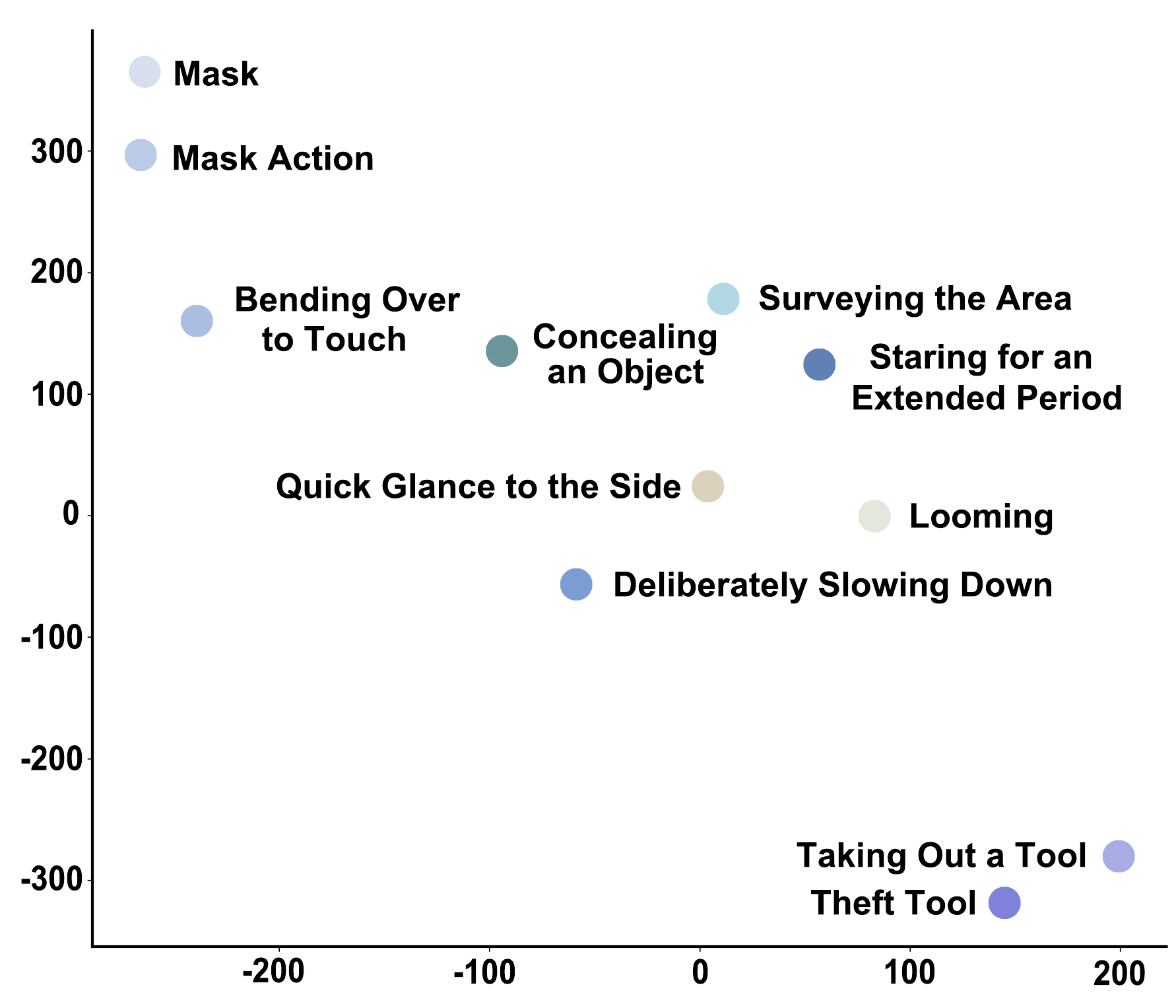}
       \caption{}
       \label{fig:tsne_visualization}
   \end{subfigure}
   \caption{Semantic space analysis of predefined suspicious action concepts: (a) similarity heatmap depicting semantic relationships between the 11 predefined concepts; (b) t-SNE visualization illustrating concept embedding distribution.}
   \label{fig:semantic_space}
\end{figure}

\subsection{Qualitative Analysis and Discussion}

Qualitative analysis of typical video clips reveals that our method accurately captures the temporal fluctuations in suspicion scores across varying scenarios. It performs well in predicting progression of suspicion scores, while maintaining a better memory of historical information in long sequences. Our method generates suspicion score curves that more closely match actual variations, while achieving 740.68 FPS, making it suitable for deployment in real-time monitoring systems.

To verify our predefined concepts' effectiveness, we also analyzed their semantic space distribution using CLIP. Fig.~\ref{fig:semantic_space} shows these 11 concepts maintain good structural properties with an average similarity of 0.7448 (std=0.0448), indicating shared semantic connections while preserving distinctiveness between action categories. Dimensionality reduction visualization confirms functionally similar actions cluster together yet remain distinguishable.

\section{Conclusion}
In this paper, we introduce the Suspicion Progression Analysis Network (SPAN), a novel method for continuous modeling of suspicion scores that captures the progressive development of hidden intentions over time. By advancing beyond the discrete classification methods prevalent in TIL tasks, SPAN enables more nuanced understanding of suspicious patterns and earlier security interventions. Our experiments demonstrate significant performance improvements with a 19.8\% MSE reduction and a 1.78\% average mAP increase over baseline methods. Notably, SPAN excels in detecting subtle, infrequent suspicious actions, achieving a 2.74\% mAP increase in low-frequency scenarios. The method's consistent performance across diverse environments—indoor/outdoor settings and varying action frequencies—confirms its robustness and generalizability. This shift from discrete classification to continuous suspicion modeling represents a significant advancement in intention understanding, equipping security systems with enhanced capabilities to detect potentially harmful intentions before they manifest as problematic actions.

\bibliographystyle{ACM-Reference-Format}
\bibliography{main_bib}

\end{document}


\title{Supplementary Materials: SPAN: Continuous Modeling of Suspicion Progression for Temporal Intention Localization}

\maketitle
\begin{table*}[t]
\centering
\caption{Ablation study on loss function components and weights. The best scores in each column are highlighted in bold. Results of our method are shaded in gray.}
\label{tab:loss_ablation}
\begin{tabular}{ccc|ccc|cccccc}
\hline
\multicolumn{3}{c|}{Loss Weights} & \multicolumn{3}{c|}{Curve Fitting} & \multicolumn{6}{c}{Temporal Localization (mAP$\uparrow$ [\%])} \\
MSE & Wave & Trend & MSE$\downarrow$ & MAE$\downarrow$ & R$^2\uparrow$ & IoU=0.3 & IoU=0.4 & IoU=0.5 & IoU=0.6 & IoU=0.7 & Avg. \\
\hline
1 & 0 & 0 & 0.013 & 0.085 & 0.025 & 10.04 & 7.53 & 4.86 & 2.19 & 0.95 & 5.11 \\
0 & 1 & 0 & 0.259 & 0.499 & -17.009 & 5.97 & 4.86 & 3.55 & 1.76 & 1.06 & 3.44 \\
0 & 0 & 1 & 0.245 & 0.523 & -18.754 & 6.07 & 4.38 & 2.73 & 1.63 & 1.01 & 3.16 \\
1 & 1 & 0 & 0.012 & 0.075 & 0.042 & 11.45 & 7.52 & 4.71 & 2.77 & 1.64 & 5.61 \\
1 & 0 & 1 & 0.007 & 0.073 & 0.034 & 10.78 & 7.66 & 4.75 & 2.22 & 0.97 & 5.28 \\
0 & 1 & 1 & 0.192 & 0.381 & -11.285 & 10.38 & 6.58 & 2.85 & 1.63 & 0.85 & 4.46 \\
1 & 1 & 1 & 0.010 & 0.058 & 0.078 & 13.18 & 8.18 & 4.97 & 3.35 & 1.45 & 6.23 \\
\cellcolor{gray!20}\textbf{1} & \cellcolor{gray!20}\textbf{0.5} & \cellcolor{gray!20}\textbf{0.3} & \cellcolor{gray!20}\textbf{0.008} & \cellcolor{gray!20}\textbf{0.059} & \cellcolor{gray!20}\textbf{0.100} & \cellcolor{gray!20}\textbf{12.08} & \cellcolor{gray!20}\textbf{9.31} & \cellcolor{gray!20}\textbf{6.72} & \cellcolor{gray!20}\textbf{4.10} & \cellcolor{gray!20}\textbf{2.43} & \cellcolor{gray!20}\textbf{6.93} \\
1 & 0.3 & 0.5 & 0.009 & 0.069 & 0.080 & 12.67 & 7.78 & 5.83 & 4.42 & 1.02 & 6.34 \\
\hline
\end{tabular}
\end{table*}

\section{Additional Ablation Studies}

\subsection{Loss Function Analysis}

We propose a \textbf{wave-aware loss} to accurately capture the dynamic characteristics of suspicion scores. This loss function consists of three key components: base loss $\mathcal{L}_{base}$, wave magnitude loss $\mathcal{L}_{magn}$, and wave trend loss $\mathcal{L}_{trend}$:
\begin{equation}
    \mathcal{L} = \mathcal{L}_{base} + \lambda_{magn}\mathcal{L}_{magn} + \lambda_{trend}\mathcal{L}_{trend}
\end{equation}

The base loss $\mathcal{L}_{base}$ employs Smooth L1 Loss to measure the disparity between predicted and ground truth values. The wave magnitude loss $\mathcal{L}_{magn}$ quantifies the deviation in first-order differences between predicted and ground truth sequences using mean squared error, capturing accurate amplitude changes in suspicion scores. The wave trend loss $\mathcal{L}_{trend}$ ensures directional consistency through direction matching, preserving the increasing or decreasing patterns in suspicion scores.

To determine the optimal weights for our multi-objective loss function, we conducted extensive ablation experiments examining various combinations of loss components. Table~\ref{tab:loss_ablation} presents the performance comparison of different loss function combinations on the HAI dataset.

Our findings demonstrate the complementary nature of the different loss components. The base MSE loss alone (1-0-0) provides reasonable performance with an average mAP of 5.11\%. When combined with other components, performance improves significantly. Through experimentation, we determined the optimal weights to be $\lambda_{magn}=0.5$ and $\lambda_{trend}=0.3$, achieving the best performance with an average mAP of 6.93\% and superior curve fitting metrics.

This weight configuration balances point-level prediction accuracy with sequence-level dynamic pattern modeling, emphasizing wave magnitude modeling slightly more than trend accuracy. This reflects our goal of ensuring both overall waveform fitting accuracy and trend prediction effectiveness. By incorporating wave magnitude loss, our model becomes sensitive to amplitude changes between consecutive time steps, while the trend loss ensures that directional changes are correctly modeled, which is essential for capturing intention transitions.

\section{Concept Definitions for Suspicious Actions}

To enable our model to better understand suspicious actions, we developed comprehensive definitions for the 11 suspicious action categories in the HAI dataset. The objective was to transform simple action labels into semantically rich definitions that capture both observable visual patterns and potential hidden intentions.

\subsection{Definition Development Process}

We employed a multi-level semantic construction method to develop definitions that balance visual observability with intentional implications:

1. We started with the 11 predefined suspicious action labels from the HAI dataset.

2. For each action label, we constructed an expanded definition using the following prompt structure:
   \begin{itemize}
      \item Request for definitions that explicitly include both visual features and potential underlying intentions
      \item Requirements for each definition to include:
         \begin{itemize}
            \item A concise general description (under 20 words)
            \item Typical visual characteristics observable in video
            \item Possible hidden intentions behind the action
         \end{itemize}
      \item Emphasis on semantic distinctiveness between similar action concepts
   \end{itemize}

3. The resulting definitions were evaluated for their computational applicability by analyzing their distribution in CLIP's semantic space. For actions with high semantic similarity, we implemented targeted refinements to enhance distinctiveness while preserving semantic relationships within the suspicious action taxonomy.

\subsection{Resulting Concept Definitions}

Through this process, we transformed the original HAI dataset action labels into the semantically enriched definitions shown in Table~\ref{tab:concept_definitions}. These definitions were incorporated into our Concept-Anchored Mapping method to establish semantic relationships between observed visual features and potential hidden intentions, enabling more effective suspicious intention localization. The effectiveness of these concept definitions is further demonstrated in Table 2 of the main paper, where our comparison between using predefined suspicious concepts (Definition) and action labels (Label) shows that our concept-based method achieves a 0.91\% higher average mAP and significantly better curve fitting metrics.

\begin{table*}[t]
\centering
\caption{Concept definitions of 11 suspicious actions in the HAI dataset.}
\label{tab:concept_definitions}
\begin{tabular}{>{\centering\arraybackslash}p{4cm}|>{\centering\arraybackslash}p{12cm}}
\hline
\textbf{Action Label} & \textbf{Expanded Definition} \\
\hline
Quick Glance to the Side & A person quickly turning their head to look to the side, often in a suspicious manner. \\
\hline
Surveying the Area & A person carefully looking around the environment to assess their surroundings. \\
\hline
Staring for an Extended Period & A person staring at an object or area for an unusually long time. \\
\hline
Bending Over to Touch & A person bending down to touch or interact with an object. \\
\hline
Looming & A person standing too close to another person or object in an intimidating manner. \\
\hline
Deliberately Slowing Down & A person noticeably slowing their pace in a suspicious way. \\
\hline
Mask & A person wearing a mask or face covering that conceals their identity. \\
\hline
Concealing an Object & A person hiding an item on their person or elsewhere to prevent others from discovering it. \\
\hline
Taking Out a Tool & A person removing a tool or implement that could be used for theft. \\
\hline
Theft Tools & A person possessing tools commonly associated with theft or break-ins. \\
\hline
Mask Action & A person putting on or adjusting a face covering or mask to hide something. \\
\hline
\end{tabular}
\end{table*}